\documentclass[Afour,sageh,times]{sagej}

\usepackage{moreverb,url}
\usepackage{pifont} % Required for \ding{55}

\usepackage[hidelinks,colorlinks,bookmarksopen,bookmarksnumbered,citecolor=blue,urlcolor=blue,allcolors=blue,pdfborder={0 0 0}]{hyperref}

\hypersetup{
    colorlinks=true, % Colored links instead of boxes
    linkcolor=blue, % Color of internal links
    citecolor=cyan, % Color of citations
    filecolor=blue, % Color of file links
    urlcolor=blue,   % Color of external links
    pdfborder={0 0 0} % Removes frame around links
}
\usepackage[dvipsnames]{xcolor}
\colorlet{linkequation}{blue}
\usepackage[colorlinks]{hyperref}
\usepackage{cleveref}

\usepackage{todonotes}
\usepackage{enumerate}
\usepackage{enumitem}
\usepackage[super]{nth}
\usepackage[amssymb]{SIunits}
\usepackage{graphicx}
\usepackage{nicefrac}
\usepackage{amssymb}
\usepackage{amsthm}
\usepackage{bm} % optional
\usepackage{footnote}
\usepackage{threeparttable}
\usepackage{epsfig}
\usepackage{graphicx}
\usepackage{balance}
\usepackage{mathptmx}
\usepackage{booktabs}
\usepackage{subcaption}
\usepackage{wrapfig}
\usepackage{caption}
\usepackage{ulem}
\usepackage{ragged2e}
\usepackage{multirow}
\usepackage{array}
\usepackage{amsmath}
\DeclareMathAlphabet{\mathcal}{OMS}{cmsy}{m}{n}
\newcolumntype{C}[1]{>{\centering\arraybackslash}p{#1}}

\usepackage{algorithm}
\usepackage{algorithmic}

\usepackage{todonotes}
\presetkeys{todonotes}{size=\tiny}{}

\newtheorem{theorem}{Theorem}%[section]

\newtheorem{lemma}[theorem]{Lemma}

\newcommand{\fref}[1]{Figure~\ref{#1}}
\newcommand{\sref}[1]{Section~\ref{#1}}
\newcommand{\tref}[1]{Table~\ref{#1}}
\newcommand{\appref}[1]{\ref{#1}}
\newcommand{\algref}[1]{Algorithm~\ref{#1}}

\newcommand{\thmref}[1]{Theorem~\ref{#1}}

\newcommand{\ie}{{i.e.},~}

\let\oldequation\equation
\let\oldendequation\endequation

\renewenvironment{equation}
  {\small\oldequation}
  {\oldendequation}

\let\oldalign\align
\let\oldendalign\endalign

\renewenvironment{align}
  {\small\oldalign}
  {\oldendalign}

\begin{document}

\runninghead{Kuan Xu et al.}

\title{GroundSLAM: A Robust Visual SLAM System for Warehouse Robots Using Ground Textures}

\author{Kuan Xu\affilnum{1}, Zheng Yang\affilnum{1}, Lihua Xie\affilnum{1}, and Chen Wang\affilnum{2}}

\affiliation{%
\affilnum{1}Nanyang Technological University, Singapore\\
\affilnum{2}University at Buffalo, USA
}

\corrauth{
Chen Wang and Lihua Xie are co-corresponding authors.\\
Chen Wang, Spatial AI \& Robotics (SAIR) Lab, Department of Computer Science and Engineering, University at Buffalo, NY 14260, USA.\\
Lihua Xie, the Centre for Advanced Robotics Technology Innovation (CARTIN), School of Electrical and Electronic Engineering, Nanyang Technological University, 50 Nanyang Avenue, Singapore 639798.}

\email{chenw@sairlab.org; elhxie@ntu.edu.sg}

\begin{abstract}
A robust visual localization and mapping system is essential for warehouse robot navigation, as cameras offer a more cost-effective alternative to LiDAR sensors. However, existing forward-facing camera systems often encounter challenges in dynamic environments and open spaces, leading to significant performance degradation during deployment. To address these limitations, a localization system utilizing a single downward-facing camera to capture ground textures presents a promising solution. Nevertheless, existing feature-based ground-texture localization methods face difficulties when operating on surfaces with sparse features or repetitive patterns.
To address this limitation, we propose GroundSLAM, a novel feature-free and ground-texture-based simultaneous localization and mapping (SLAM) system. GroundSLAM consists of three components: feature-free visual odometry, ground-texture-based loop detection and map optimization, and map reuse. Specifically, we introduce a kernel cross-correlator (KCC) for image-level pose tracking, loop detection, and map reuse to improve localization accuracy and robustness, and incorporate adaptive pruning strategies to enhance efficiency.
Due to these specific designs, GroundSLAM is able to deliver efficient and stable localization across various ground surfaces such as those with sparse features and repetitive patterns.
To advance research in this area, we introduce the first ground-texture dataset with precise ground-truth poses, consisting of 131k images collected from 10 kinds of indoor and outdoor ground surfaces. Extensive experimental results show that GroundSLAM outperforms state-of-the-art methods for both indoor and outdoor localization. We release our code and dataset at \url{https://github.com/sair-lab/GroundSLAM}.

\end{abstract}


\maketitle

\section{Introduction} \label{sec:introduction}

Warehouse automation relies heavily on accurate and robust simultaneous localization and mapping (SLAM) systems to enable efficient robot navigation \citep{wu2023rf}. Among the available sensing technologies, cameras provide a cost-effective alternative to LiDAR sensors, making the visual SLAM system an attractive choice for warehouse robotics \citep{kazerouni2022survey}. As a result, significant research efforts have been dedicated to developing visual SLAM systems for warehouse robots \citep{wilhelm2024lightweight}.

\begin{figure}[t]
    % \vspace{0.em}
    \centering
    \includegraphics[width= 0.99\linewidth]{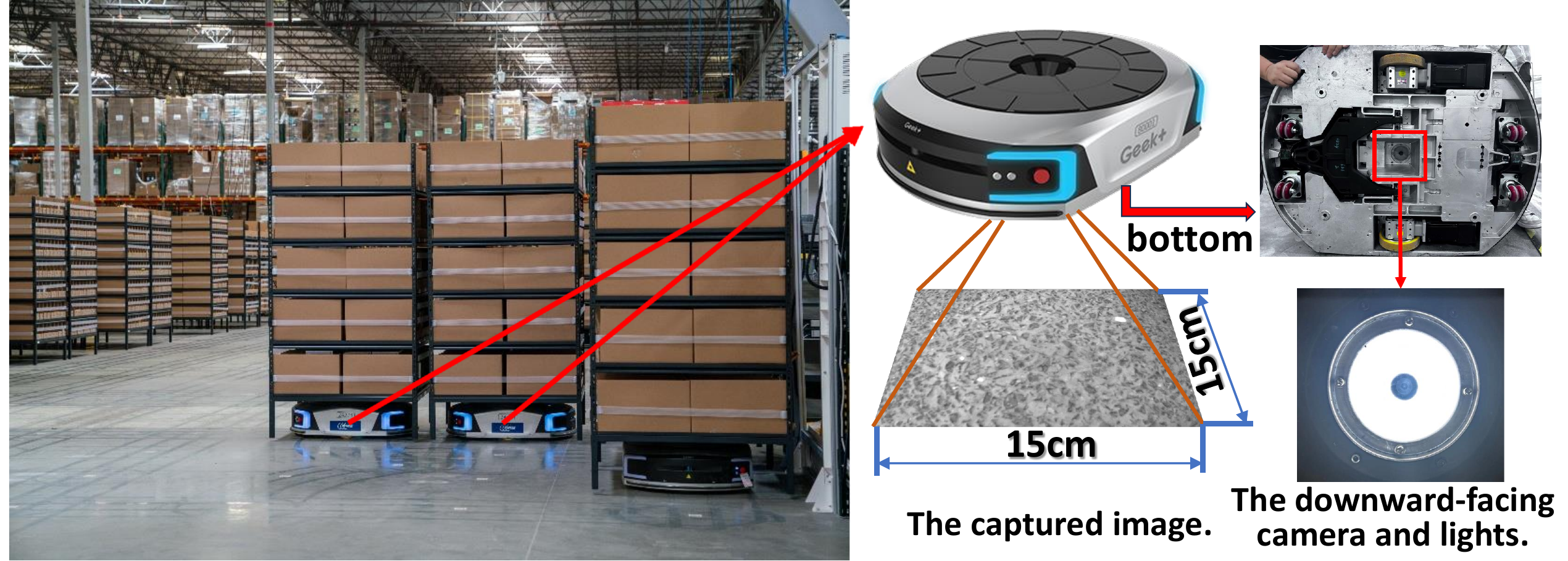}
    \caption{In the warehouse, dynamic objects (robots and storage racks) and distant features make the localization 
    with a forward-facing camera or LiDAR very challenging.}
    \label{fig:warehouse}
    \vspace{-0.5em}
\end{figure}

Despite significant advancements, most visual SLAM systems are developed for forward-facing cameras, which often struggle in complex environments with dynamic objects, large open spaces, occlusions, motion blur, and the lack of stable visual landmarks in textureless areas \citep{xu2022bifocal, qiu2022airdos}.
For instance, in environments illustrated in \fref{fig:warehouse}, where multiple robots collaborate to transport goods, the highly dynamic nature often leads to significant localization errors. Moreover, warehouses are typically vast, causing static features to be distant from the camera and reducing the localization accuracy. 
As a result, tag-based localization \citep{moura2021graph} remains the dominant approach for warehouse applications. This involves placing a multitude of QR codes on the floor and utilizing a camera positioned downward to detect them for localization purposes. However, accurately placing these codes is time-consuming, often taking weeks or months. Besides, additional sensors, such as IMUs and wheel odometry, are still needed for localization between consecutive QR codes. Any drift error in these sensors can cause failures in detecting the next code, especially when the camera is close to the ground and has a narrow field of view.

To enhance the reliability and adaptability of localization, several systems have chosen to adopt ground texture-based localization using downward-facing cameras \citep{kozak2016ranger, chen2018streetmap, nakashima2019akaze, zhang2019high, schmid2020ground, hart2023monocular, wilhelm2024lightweight}.
These systems detect ground image features, which requires point-level feature matching to determine the robots' pose.
However, point-level matching relies on the availability of an adequate number of distinctive features, which may not always be guaranteed, particularly when dealing with ground textures with repetitive patterns or a scarcity of identifiable corners.
Therefore, there is a pressing necessity for the development of strategies that go beyond point-level matching to enhance the robustness of ground-texture-based visual SLAM system for warehouse robots.

Motivated by these observations, we argue that ground-texture-based visual SLAM systems can benefit from image-level matching.
This is because (1) image-level matching is not dependent on the presence of frequent corners or textures, making it robust against low-texture environments and image blurriness and (2) image-level matching can seamlessly connect dominant patterns using comprehensive global information, preventing the system from losing track on the ground when dealing with repetitive local patterns.
Despite those appealing properties, a direct implementation of image matching for ground-texture-based visual SLAM still presents significant challenges.
This is because most existing image-level matching in visual SLAM systems requires RGB-D inputs \citep{Newcombe:2011gw, whelan2015elasticfusion, Newcombe:2015hz} or heavy learning process \citep{wang2021tartanvo, fu2024islam, murai2024mast3r}. RGB-D-based methods heavily rely on iterative closest point (ICP)-based methods and thus cannot estimate the relative transformation between two planes. Deep-learning-based methods require a large number of labeled images, which poses significant challenges for ground-texture-based localization. Besides, they also demand substantial computational resources to run in real time, and are thus difficult to deploy on warehouse robots with low power consumption and cost requirements.

To address these challenges, we propose GroundSLAM, a novel visual SLAM system designed for warehouse robots. GroundSLAM employs a downward-facing camera to capture ground textures for localization. To enhance resilience against low-texture surfaces and repetitive local patterns, we extend our previous work, a kernel cross-correlator (KCC) \citep{wang2018kernel}, for image-level pose tracking, loop detection, and map reuse, and develop pruning strategies to enhance the efficiency of the overall system.
Furthermore, we develop efficient modules for map storage, loading, and reuse, enabling multiple robots to achieve drift-free localization within a unified map.
These designs make GroundSLAM a highly cost-effective solution, well-suited for warehouse environments where affordability, robustness, and real-time operation are critical.
Additionally, KCC provides an efficient closed-form solution, ensuring both efficiency and real-time performance with minimal computational overhead.
However, we identified that the original formulation of KCC lacked a rigorous theoretical proof. To address this gap, we provide a more detailed and formal proof of KCC in the Fourier domain, strengthening its theoretical foundation and ensuring its validity for GroundSLAM.
In summary, our contributions include
\begin{itemize}[noitemsep,topsep=0pt]
  \item We propose a novel complete visual SLAM pipeline for warehouse robots using ground textures, which includes visual odometry, loop closure detection, and map reuse. Our system can provide robust pose estimation and localization in environments with many dynamic objects or open spaces, such as warehouses.
  \item We present a rigorous and detailed proof for KCC and introduce it to the ground texture-based visual SLAM system. We show that KCC can perform efficient image-level matching for both visual odometry and loop closure detection, achieving significantly more robust performance than feature-based methods in environments with low texture and repetitive corners. 
  \item We collect and release a ground texture dataset for ground-texture-based localization which contains 131k images collected from 10 different kinds of surfaces. As far as we know, it is the first publicly available ground texture dataset with accurate ground truth. Additionally, we perform extensive experiments that prove the efficiency and effectiveness of GroundSLAM. The results show that our GroundSLAM outperforms the state-of-the-art (SOTA) systems on various ground textures. We release the source code and data at \url{https://github.com/sair-lab/GroundSLAM} to benefit the community.
\end{itemize}

The remainder of this article is organized as follows. In \sref{sec:related-work}, we review the related works on the ground-texture-based localization, image-level matching methods and the correlation filter. In \sref{sec:methodology}, we present the kernel cross-correlator with a detailed proof to estimate the relative transformation between two images. Based on KCC, a ground-texture-based visual SLAM system is designed in \sref{sec:system_architecture}. We present the experimental results in \sref{sec:experiments} to verify the robustness of GroundSLAM compared with the SOTA systems. This article is concluded in \sref{sec:conclusion}.

\section{Related Works} \label{sec:related-work}

\subsection{Ground-Texture-Based Localization}
% \subsubsection{Feature-Based Methods}
\paragraph{Feature-Based}
Most ground-texture-based visual localization systems utilized feature detection and matching to perform the iterative 3-DOF pose estimation. Ranger \citep{kozak2016ranger} detected CenSurE \citep{agrawal2008censure} features on the road surfaces and matches them using ORB \citep{Anonymous:2011ug} descriptors. Then a coarse homography is computed with RANSAC to discard false matches. Finally, a fine 2D pose is estimated using the remaining correspondences. StreepMap \citep{chen2018streetmap} proposed two algorithms and names them feature-based approach and line-based approach, respectively. The former approach detects SURF \citep{bay2006surf} features and tracks them with Lucas-Kanade method \citep{shi1994good}. The latter approach fused the measurements from line features and an IMU together using an Extended Kalman Filter (EKF). Micro-GPS \citep{zhang2019high} used the SIFT scale-space DoG detector and gradient orientation histogram descriptor \citep{lowe2004distinctive} for both mapping and localization.

\cite{schmid2020ground} improved Micro-GPS by proposing identity feature matching, where only identical descriptors are considered as matches. Then they extend this work by replacing SIFT feature detection with randomly sampled points \citep{schmid2020ground}. \cite{zhang2018learning} designed a Convolutional Neural Network (CNN) to detect feature points on the ground texture image. \cite{hart2023monocular} proposed ground texture SLAM (GT-SLAM), where ORB features are detected and matched using Fast Library for Approximate Nearest Neighbors (FLANN) method \citep{muja2009fast}. A factor graph is constructed and optimized to estimate the transformation between adjacent frames. To correct the drift error, they train a vocabulary tree using the Bag of Words (BoW) library \citep{galvez2012bags} to find loop closures. \cite{wilhelm2024lightweight} proposed a lightweight ground-texture-based global localization system that supports both ORB and SIFT features.
As far as we know, GT-SLAM \citep{hart2023monocular} is the latest state-of-the-art ground-texture-based localization system that has been open-sourced, so we take it as the most important baseline in the experiments.

% \subsubsection{Correction-Based Methods}
\paragraph{Correction-Based}
\cite{4337964} analyzed the eigenvalues of a matrix composed of image intensity gradients from the input image, aiming to detect areas with bidirectional texture. Then small patches ($25 \times 25$ pixel) around these points are selected as features and matched using cross-correlation. \cite{zaman2007high} estimated the transformation between two images by finding the maximum peak of their cross-correlations. The system needs an accurate prior rotation with the error within $2^{\circ}$. Then the second image is transformed with a set of rotations and translations to compute their cross-correlations. The transformation with the maximum peak is regarded as the true one. Similar methods were also used in \cite{nourani2011correlation} and \cite{ZAMAN201982}. These two systems employ template matching and cross-correlation to estimate the 3-DOF relative pose between two frames. \cite{7354239} detected two groups of points for each image, and then used a conventional template matching to find the correction of points on different frames. 
\cite{birem2018visual} utilized Fourier transform and phase correction to estimate the relative transformation between two ground texture images in a visual odometry system.
\cite{xu2021rethinking} proposed a VO system for downward-looking cameras on UAVs. They extended the Fourier-Mellin transform to make it applicable to multi-depth scenes. 
Subsequently, they refined their system by optimizing its efficiency and incorporating a back-end optimization module \citep{jiang2023optimizing}.
Unlike these methods, we introduce a kernel cross-correlator to estimate the image transformation, enhancing the robustness of our SLAM system.

% \subsubsection{Datasets}
\paragraph{Datasets}
To the best of our knowledge, only two ground texture datasets for localization are publicly available. One is the Micro-GPS dataset \citep{zhang2019high} and the other is the HD Ground dataset \citep{schmid2022hd}. The authors of the Micro-GPS dataset collected data on 7 kinds of ground textures. Their imaging system consists of a Point Grey CM3 grayscale camera pointed downwards at the ground and surrounded by a set of LED lights. For each kind of texture, both databases for mapping and test sequences for localization were collected. The HD Ground dataset contains 11 kinds of ground textures. Similarly, the data is collected by a ground-facing camera. The recording area is shielded from external lighting and illuminated by a 24V, 72Watt LED ring. However, the ground truths of these two datasets are both from image stitching instead of high-precision pose measurement devices, so their accuracy highly depends on the performance of feature detection and matching of the image-stitching system.
In contrast, our dataset includes ground-truth data obtained from a Leica Nova MS60 MultiStation laser tracker, making it more suitable for evaluating SLAM systems.

\subsection{Image-Level Matching} \label{sec:related-matching}
Image-level matching for pose estimation can be categorized into ICP-based, learning-based, and correction-based approaches. ICP-based methods \citep{Newcombe:2011gw, whelan2015elasticfusion, Newcombe:2015hz} rely on RGB-D input, making them unsuitable for ground-texture localization, where a single monocular camera is used to estimate the motion between two planes.
Learning-based algorithms \citep{wang2017deepvo, bloesch2018codeslam, wang2021tartanvo, shen2023dytanvo, fu2024islam, murai2024mast3r} utilize deep neural networks to directly or indirectly estimate the pose between two images, achieving outstanding performance in visual SLAM systems with forward-facing cameras. They typically require a large amount of labeled data. However, publicly available ground texture datasets for localization are limited, and none of them provide accurate ground truth. Additionally, learning-based methods require GPUs to run in real time, making their cost prohibitive for deployment on warehouse robots. Correction-based approaches \citep{kazik2011visual, bulow2018scale, xu2021rethinking, jiang2023optimizing} utilize the Fourier-Mellin transform to estimate relative rotation and translation. Our system is also correction-based. However, unlike the approaches mentioned above, we introduce kernel methods to enhance the system's robustness.

\subsection{Correlation Filter} \label{sec:related-correlation}

Correlation filter is a class of classifier, which is specifically optimized to produce sharp peaks in the output to achieve accurate translation estimation \citep{boddeti2012advances}.
By specifying the desired response at every location, the average synthetic exact filter (ASEF) generalizes across the entire training set by averaging multiple exact filters \citep{Bolme:2009jq}.
To overcome the overfitting problem of ASEF, the minimum output sum of squared error (MOSSE) filter adds a regularization term and introduces it into visual tracking \citep{Anonymous:2010tm}. 
Due to its complexity of $\mathcal{O}(n\log n)$ ($n$ is the number of pixels), MOSSE has a superior speed and thus ignited the boom and development of CF-based tracking.
Kernelized correlation filter (KCF) brings circulant training structure into kernel ridge regression \citep{Henriques:2015jy}. This enables learning with element-wise operation instead of costly matrix inversion, providing much more robustness while still with reasonable learning speed.
Multi-kernel correlation filter (MKCF) \citep{Tang:2015jxba} extended KCF to multiple kernels, which further improves the accuracy.
To alleviate the boundary effect of CFs, zero aliasing correlation filter (ZACF) \citep{Fernandez:2015fi} introduced the zero-aliasing constraints and provides both closed-form and iterative proximal solutions by ensuring that the optimization criterion for a given CF corresponds to a linear correlation rather than a circular correlation. However, it requires heavy computation and is not suitable for real-time applications.
Discriminative scale space tracking (DSST) \citep{Danelljan:2017ij} was proposed to learn multiple MOSSE on different scales, enabling estimation of both translation and scale at the cost of repeated calculations of MOSSE.
STRCF \citep{li2018learning} introduced temporal regularization to spatially regularized correlation filters.
Spatially local response map variation was introduced by \cite{li2020autotrack} as spatial regularization to make the correlation filter focus on the trustworthy object parts.
Due to the robustness and efficiency, we propose to introduce KCC into all the components of GroundSLAM, including pose tracking, loop detection, and map reuse, resulting in an extremely robust visual SLAM system for warehouse robots.

\section{Transformation Estimation} \label{sec:methodology}

The core of GroundSLAM is to estimate the relative transformation (movements) between two images, including translation and rotation.
Therefore, prior to delving into the system architecture, we present the kernel cross-correlator for image transformation estimation to provide a clearer understanding.
Given a keyframe $\mathbf{x}\in\mathbb{R}^{M\times N}$ and a current image $\mathbf{z}\in\mathbb{R}^{M\times N}$, our objective is to find a transformation $\mathcal{T}:\mathbb{R}^{M\times N}\mapsto\in\mathbb{R}^{M\times N}$, so that the transformed keyframe $\mathcal{T}(\mathbf{z})$ has the maximum kernel similarity with the current image $\mathbf{x}$:
\begin{equation}\label{eq:objective}
  \mathcal{T}^\star = \arg_{\mathcal{T}}\max \boldsymbol{\kappa}\left(\mathbf{x},  \mathcal{T}(\mathbf{z})\right)
\end{equation}
where $\boldsymbol{\kappa}:\mathbb{R}^{M\times N}\times\mathbb{R}^{M\times N}\mapsto\mathbb{R}^+$ is a robust kernel function to measure the similarity of two signals. In this paper, the image transformation $\mathcal{T}$ can be translation, rotation, or both.

A common approach to solving the objective function \eqref{eq:objective} is to optimize the kernel function with respect to the images using \textit{gradient descent}. However, this method is often computationally expensive and prone to convergence issues, as \eqref{eq:objective} contains multiple local minima, making the solution highly sensitive to initialization. We next show that the objective function \eqref{eq:objective} can be easily solved via our kernel cross-correlator in a computational complexity of $\mathcal{O}(n\log n)$, where $n=MN$ is the number of pixels.

\subsection{Kernel Cross-Correlation}

Our fundamental insight is that if we can compute the kernel functions for every potential transformation $\mathcal{T}_i$ within $\mathcal{T}$, then the solution to \eqref{eq:objective} is the transformation corresponding to the highest kernel function value. In this context, we have
\begin{equation}\label{eq:kernel-objective-any}
\mathcal{T}^\star
=\arg_{\mathcal{T}}\max\left(\kappa(\mathbf{x}, \mathcal{T}_0(\mathbf{z})), \cdots, \kappa(\mathbf{x}, \mathcal{T}_{m-1}(\mathbf{z})\right),
\end{equation}
where $\mathcal{T}_i(\mathbf{z})$ is a transformation of $\mathbf{z}$. 
For the sake of simplicity, we denote images as vectorized signals $\mathbf{z,x}\in\mathbb{R}^n$, where $n=MN$. The following conclusions can be easily extended to matrices.
Since an actual transformation is continuous, we can sample it uniformly to obtain $\{{\mathcal{T}_0, ..., \mathcal{T}_{m-1}}\}$, where the relative transformation between any adjacent ${\mathcal{T}}_i$, \ie $\{\mathcal{T}_k, \mathcal{T}_{k+1}\}, k \in [0, m-2]$, is equal.
Note that the index starts from 0 to denote no transformation, \ie $\mathcal{T}_0(\mathbf{z})=\mathbf{z}$.
Define the kernel vector as $\boldsymbol{\kappa}_{\mathbf{z}}(\mathbf{x}) = [\kappa(\mathbf{x}, \mathcal{T}_0(\mathbf{z})), \cdots, \kappa(\mathbf{x}, \mathcal{T}_{m-1}(\mathbf{z})]^T$, then our kernel cross-correlation (KCC) is defined as:
\begin{equation}\label{eq:kernel-correlation-time}
	\mathbf{g}(\mathbf{x}) =  \boldsymbol{\kappa}_{\mathbf{z}}(\mathbf{x}) \otimes \mathbf{h},
\end{equation}
where $\otimes$ is the linear circular cross-correlation operator \citep{bracewell1986fourier},
and $\mathbf{h} \in\mathbb{C}^{m}$ is an unknown correlator to be estimated,  and $\mathbf{g}\in\mathbb{C}^{m}$ is the correlation output.
Due to the cross-correlation theorem \citep{bracewell1986fourier}, KCC \eqref{eq:kernel-correlation-time} can be calculated in frequency domain:
\begin{equation}\label{eq:kernel-correlation}
	\hat{\mathbf{g}}(\mathbf{x}) =  \hat{\boldsymbol{\kappa}}_{\mathbf{z}}(\mathbf{x}) \odot \hat{\mathbf{h}}^*,
\end{equation}
where $\odot$ is the element-wise multiplication, $\cdot^*$ is the complex conjugate, and $\hat{\cdot}$ denotes the fast Fourier transform (FFT) $\mathcal{F}(\cdot)$.
The background of the linear cross-correlation and cross-correlation theorem can be found in \appref{app:appendix-cross-correlation}.

Before delving further, we will first define \textit{equivariance} for clearer comprehension and then demonstrate that KCC is equivariant to any affine transformations, which is an important property for pose estimation in GroundSLAM.

\begin{lemma}[Equivariance \citep{gaunce2006equivariant}]
Consider a function $\mathbf{f}$ and a transformation $\mathcal{T}$, the function $\mathbf{f}$ is equivariant to $\mathcal{T}$ if
\begin{equation}
	\mathbf{f}(\mathcal{T}(\mathbf{x}))=\mathcal{T}'(\mathbf{f}(\mathbf{x}))
\end{equation}
for all input $\mathbf{x}$, where $\mathcal{T}'$ might be the same as $\mathcal{T}$ or another related transformation. In simple terms, a function is said to be equivariant if the input changes in a certain way, the output changes in a predictable and corresponding manner.
\end{lemma}

\begin{theorem}[KCC Equivariance]\label{thm:kcc-equivariance}
Kernel cross-correlator in \eqref{eq:kernel-correlation-time} is equivariant to transformation $\mathcal{T}_i$, if $\mathcal{T}_i$ is periodical: $\mathcal{T}_i(\mathbf{x})=\mathcal{T}_{i+m}(\mathbf{x})$, where $m$ is the period. In other words,
	\begin{equation}
		\mathbf{g}(\mathcal{T}_{i}(\mathbf{x}))=\boldsymbol{\kappa}_{\mathbf{z}}(\mathcal{T}_{i}(\mathbf{x})) \otimes \mathbf{h} = (\boldsymbol{\kappa}_{\mathbf{z}}(\mathbf{x}) \otimes \mathbf{h})_{(i)}=\mathbf{g}_{(i)}(\mathbf{x}),
	\end{equation}
  where $\cdot_{(i)}$ denotes the vector circularly shifted by $i$ elements.
\end{theorem}

\begin{proof}
  Since circular translation is a periodical transform function and cross-correlation is equivariant to circular translation \citep{gaunce2006equivariant}, we have $(\mathbf{a}\otimes\mathbf{b})_{(i)} = \mathbf{a}_{(i)}\otimes\mathbf{b}$, where $\mathbf{a},\mathbf{b}$ are two random vectors.
	Assume $\mathbf{g}'=\boldsymbol{\kappa}_{\mathbf{z}}(\mathcal{T}_{j}(\mathbf{x})) \otimes \mathbf{h}$, then
	\begin{equation}
	\mathbf{g}'= [\kappa(\mathcal{T}_{j}(\mathbf{x}), \mathcal{T}_{0}(\mathbf{z})), \cdots, \kappa(\mathcal{T}_{j}(\mathbf{x}), \mathcal{T}_{m-1}(\mathbf{z}))]\otimes\mathbf{h}.
	\end{equation}
Since $\mathcal{T}_{j}$ is periodical, we have
  \begin{equation}
    \boldsymbol{\kappa}_{\mathbf{z}}(\mathcal{T}_{j}(\mathbf{x})) = (\boldsymbol{\kappa}_{\mathbf{z}}(\mathbf{x}))_{(j)}.
  \end{equation}
  This means that
	\begin{equation}
		\boldsymbol{\kappa}_{\mathbf{z}}(\mathcal{T}_{j}(\mathbf{z})) \otimes \mathbf{h} = (\boldsymbol{\kappa}_{\mathbf{z}}(\mathbf{z}))_{(j)} \otimes \mathbf{h} = (\boldsymbol{\kappa}_{\mathbf{z}}(\mathbf{z}) \otimes \mathbf{h} )_{(j)},
	\end{equation}
	which completes the proof.
\end{proof}

In the proof above, we demonstrate that the transformation $\mathcal{T}$ should ideally be periodic, as in the case of rotation. However, this periodicity is not always a practical requirement. For instance, image translation does not inherently follow circular translation, resulting in a “boundary effect” \citep{Anonymous:2010tm}. In our experiments, we adopt a simple yet effective strategy, doubling the signal length, to mitigate this issue. This approach helps alleviate boundary effects when applying other transformations such as translation and scaling. For the sake of clarity and simplicity, we provide its details in \appref{app:appendix-boundary}.

\vspace{0.2em}
\subsection{The Closed-form Solution to KCC}

The basic idea to estimate image transformation using KCC is that we can take advantage of the equivariance property to convert the effect of a transformation into the translation of its output.
Concretely, if we can find a correlation filter $\mathbf{h}^{\star}$, that is able to map the kernel vector $\boldsymbol{\kappa}_{\mathbf{z}}(\mathbf{z})$ to a predefined correlation output $\mathbf{g}^{\star} := \boldsymbol{\kappa}_{\mathbf{z}}(\mathbf{z}) \otimes \mathbf{h}^{\star}$, then the best transformation of $\mathbf{x}$ denoted as $ \mathcal{T}^\star(\mathbf{x})$ is corresponding to the translation of $\boldsymbol{\kappa}_{\mathbf{z}}(\mathbf{x}) \otimes \mathbf{h}^\star$ relative to the predefined target $\mathbf{g}^{\star}$.

For simplify, we can set $\mathbf{g}^\star$ as a single peak vector, then the translation of $\boldsymbol{\kappa}_{\mathbf{z}}(\mathbf{x}) \otimes \mathbf{h}^\star$ relative to $\mathbf{g}^\star$ can be found by calculating the translation between their maximum values:
\begin{subequations}\label{eq:prediction}
\begin{align}
    i^\star &= \arg_{i}\max \mathbf{g}^{\star}[i], \\
    j^\star &= \arg_{j}\max \left(\boldsymbol{\kappa}_{\mathbf{z}}(\mathbf{x}) \otimes \mathbf{h}^\star\right)[j], \\
    \mathcal{T}^\star &= \mathcal{T}_{(j^\star-i^\star)\%m}, 
\end{align}
\end{subequations}
where $\%$ is the modulo operator.

In this way, the original problem becomes how to find the correlator $\mathbf{h}^\star$ satisfying $\mathbf{g}^{\star} = \boldsymbol{\kappa}_{\mathbf{z}}(\mathbf{z}) \otimes \mathbf{h}^{\star}$. Intuitively, it can be solved by finding the optimal solution to the function:
\begin{equation}\label{eq:objective-time}
  \min_{\mathbf{h}}\|\boldsymbol{\kappa}_{\mathbf{z}}(\mathbf{z}) \otimes \mathbf{h} - \mathbf{g}^{\star}\|^2 + \lambda \|\mathbf{h}\|^2,
\end{equation}
where the second term is a regularization to prevent overfitting. 
However, objective \eqref{eq:objective-time} is difficult to solve and has high computational complexity. Inspired by the fact that cross-correlation can be accelerated in the frequency domain, we instead minimize its alternative in the frequency domain to take advantage of the efficient element-wise operation:
\begin{equation}\label{eq:kernel-objective}
\min_{\hat{\mathbf{h}}^*}\|\hat{\boldsymbol{\kappa}}_{\mathbf{z}}(\mathbf{z}) \odot \hat{\mathbf{h}}^* - \hat{\mathbf{g}}^\star \|^2 + \lambda \|\hat{\mathbf{h}}^*\|^2,
\end{equation}
where $\hat{\mathbf{g}}^\star := \mathcal{F}(\mathbf{g}^\star)$ and $\hat{\mathbf{h}}^* := \mathcal{F}^*(\mathbf{h})$ are the corresponding FFT of $\mathbf{g}$ and $\mathbf{h}$ to simplify the notations.

We adopt the objective function \eqref{eq:kernel-objective} instead of \eqref{eq:objective-time} as \eqref{eq:kernel-objective} allows for an efficient closed-form solution in the frequency domain, which is critical for ensuring computational efficiency and real-time performance.

\newcommand{\kvi}{\hat{\boldsymbol{\kappa}}_{\mathbf{z}}(\mathbf{z})}
\newcommand{\kvic}{\hat{\boldsymbol{\kappa}}^*_{\mathbf{z}}(\mathbf{z})}
\begin{theorem}[KCC]\label{thm:kcc}
  There is a closed-form solution to the objective function \eqref{eq:kernel-objective} for kernel cross-correlator \eqref{eq:kernel-correlation}, 
	\begin{equation}\label{eq:kernel-solution}
	\hat{\mathbf{h}}^* = \frac{\hat{\mathbf{g}}^\star \odot \kvic}{\kvic \odot \kvi +\lambda},
	\end{equation}
	where the operator $\frac{\cdot}{\cdot}$ denotes the element-wise division and $\hat{\mathbf{h}}^*$ is the conjugate of Fourier transform of $\mathbf{h}^\star$.
\end{theorem}

\begin{proof}
  Without ambiguity, we will denote $\mathbf{g}^\star$ as $\mathbf{g}$ in this proof for simplicity.
  To solve the optimization problem \eqref{eq:kernel-objective},  we set its first derivative with respect to $\hat{\mathbf{h}}^*$ to zero, i.e.,
  \begin{equation}\label{eq:kernel-derivative}
  \frac{\partial }{\partial \hat{\mathbf{h}}^*}\left( \|\hat{\boldsymbol{\kappa}}_{\mathbf{z}}(\mathbf{z}) \odot \hat{\mathbf{h}}^* - \hat{\mathbf{g}} \|^2 + \lambda \|\hat{\mathbf{h}}^*\|^2 \right) = 0.
  \end{equation}
  Then by calculating the square ($\|a\|^2=aa^*$), we have
  \begin{equation}
   \begin{aligned}
  \frac{\partial }{\partial \hat{\mathbf{h}}^*}\Bigg(
  \kvi\odot\hat{\mathbf{h}}^*\odot\kvic\odot\hat{\mathbf{h}}-
  \kvi\odot\hat{\mathbf{h}}^*\odot\hat{\mathbf{g}}^* &\\
  -\hat{\mathbf{g}}\odot\kvic\odot\hat{\mathbf{h}}+
  \hat{\mathbf{g}}\odot\hat{\mathbf{g}}^*+\lambda \hat{\mathbf{h}}^* \odot \hat{\mathbf{h}}\Bigg)=0. &
  \end{aligned}
  \end{equation}
  We next obtain the derivative w.r.t. $\hat{\mathbf{h}}^*$ by taking $\hat{\mathbf{h}}$ as an independent variable of $\hat{\mathbf{h}}^*$ \citep{messerschmitt2006stationary}, thus
  \begin{equation}
  0=\kvi\odot\kvic\odot\hat{\mathbf{h}}-\kvi\odot\hat{\mathbf{g}}^* + \lambda \hat{\mathbf{h}}.
  \end{equation}
  Therefore, we can obtain $\hat{\mathbf{h}}$ as
  \begin{equation}\label{eq:kernel-solution-corollary}
    \hat{\mathbf{h}} = \frac{\kvi \odot \hat{\mathbf{g}}^{*}}{\hat{\boldsymbol{\kappa}}_{\mathbf{z}}(\mathbf{z}) \odot \hat{\boldsymbol{\kappa}}_{\mathbf{z}}^*(\mathbf{z}) +\lambda}.
  \end{equation}
  Due to the properties of the complex conjugate, i.e., $a^*\cdot b^*=(a\cdot b)^*$, $a^*/b^* =(a/b)^*$, and $a^* + b^*=(a+b)^*$, the solution \eqref{eq:kernel-solution} can be obtained from \eqref{eq:kernel-solution-corollary} by taking the conjugate on both sides, which completes the proof.
\end{proof}
\thmref{thm:kcc} indicates that we can calculate a correlator $\hat{\mathbf{h}}^*$ for every keyframe, and then predict the transformation of any image $\mathbf{x}$ relative to the keyframe via \eqref{eq:prediction}. This can be used for both visual odometry and loop closure detection.

In practice, the solution \eqref{eq:kernel-solution} can be further simplified. We can add a weighted regularization to the objective function \eqref{eq:kernel-objective-2}, which results in a weighted regularized KCC:
\begin{equation}\label{eq:kernel-objective-2}
\min_{\hat{\mathbf{h}}^*}\|\hat{\boldsymbol{\kappa}}_{\mathbf{z}}(\mathbf{z}) \odot \hat{\mathbf{h}}^* - \hat{\mathbf{g}} \|^2 + \lambda\|\sqrt{\hat{\boldsymbol{\kappa}}_{\mathbf{z}}(\mathbf{z})}\odot\mathbf{h}\|^2.
\end{equation}

\begin{theorem}[Weighted Regularized KCC]\label{thm:kcc-simple-solution}
  There is a closed-form solution to the objective  \eqref{eq:kernel-objective-2} for KCC \eqref{eq:kernel-correlation}: \newcommand{\kv}{\hat{\boldsymbol{\kappa}}_{\mathbf{z}}(\mathbf{z})}
	\begin{equation}\label{eq:kcc-solution}
	\hat{\mathbf{h}}^* = \frac{\hat{\mathbf{g}}^\star} {\kv+\lambda}.
	\end{equation}
\end{theorem}

\begin{proof}
  Since only the regularization of the objective function is changed, the solution to \eqref{eq:kernel-objective-2} can be obtained by updating the regularization term of \eqref{eq:kernel-solution-corollary}, hence we have
  \newcommand{\kvc}{\hat{\boldsymbol{\kappa}}^*_{\mathbf{z}}(\mathbf{z})}
  \begin{equation} \label{eq:kcc-solution-proof}
  \begin{aligned}
    \hat{\mathbf{h}} &= \frac{\hat{\mathbf{g}}^{*} \odot \kvi}{\hat{\boldsymbol{\kappa}}_{\mathbf{z}}(\mathbf{z}) \odot \hat{\boldsymbol{\kappa}}_{\mathbf{z}}^*(\mathbf{z}) +\lambda\kvi},\\
    &=\frac{\hat{\mathbf{g}}^*} {\kvc+\lambda}.
  \end{aligned}
  \end{equation}
  Similarly, solution \eqref{eq:kcc-solution} can be obtained by taking the conjugate on both sides, which completes the proof.
\end{proof}

Therefore, we have the following theorem to estimate the transformation between two signals by setting the correlation target as a single peak binary vector at the first element.

\begin{theorem}\label{thm:transform}
  The transformation between two signals $\mathbf{x},\mathbf{z}\in\mathbb{R}^{n}$ can be estimated in terms of the minimum squared spectrum error \eqref{eq:kernel-objective-2}, i.e., $\mathbf{x}=\mathcal{T}^\star(\mathbf{z})$ via KCC as
  \newcommand{\kv}{\hat{\boldsymbol{\kappa}}_{\mathbf{z}}(\mathbf{z})}
  \newcommand{\kx}{\hat{\boldsymbol{\kappa}}_{\mathbf{z}}(\mathbf{x})}
  \begin{equation}
    \mathcal{T}^\star = \mathcal{T}_{j^\star},
  \end{equation}
  where
  \begin{equation}\label{eq:maximum-transform} 
      j^\star = \arg_{j}\max \mathcal{F}^{-1}[j]\left(\frac{\hat{\mathbf{g}}^\star \odot \kx} {\kv+\lambda}\right),
  \end{equation}
  if $\mathbf{g}^\star$ is the correlation target and is predefined as
\begin{equation}\label{eq:single-peak}
    \mathbf{g}^\star[i]= 
    \begin{cases} 
        1 & i=0 \\ 
        0 & i=1,2,\cdots, m-1
    \end{cases}.
\end{equation}
\end{theorem}
\begin{proof}
  Substitute \eqref{eq:kcc-solution} into \eqref{eq:kernel-correlation}, then we can obtain \eqref{eq:maximum-transform} by taking $i^\star=0$ in \eqref{eq:prediction} because of $0=\arg_i\max \mathbf{g}^\star[i]$ in \eqref{eq:single-peak}. 
\end{proof}

\begin{figure*}[t]
    \vspace{0.5em}
    \centering
    \includegraphics[width=0.99\linewidth]{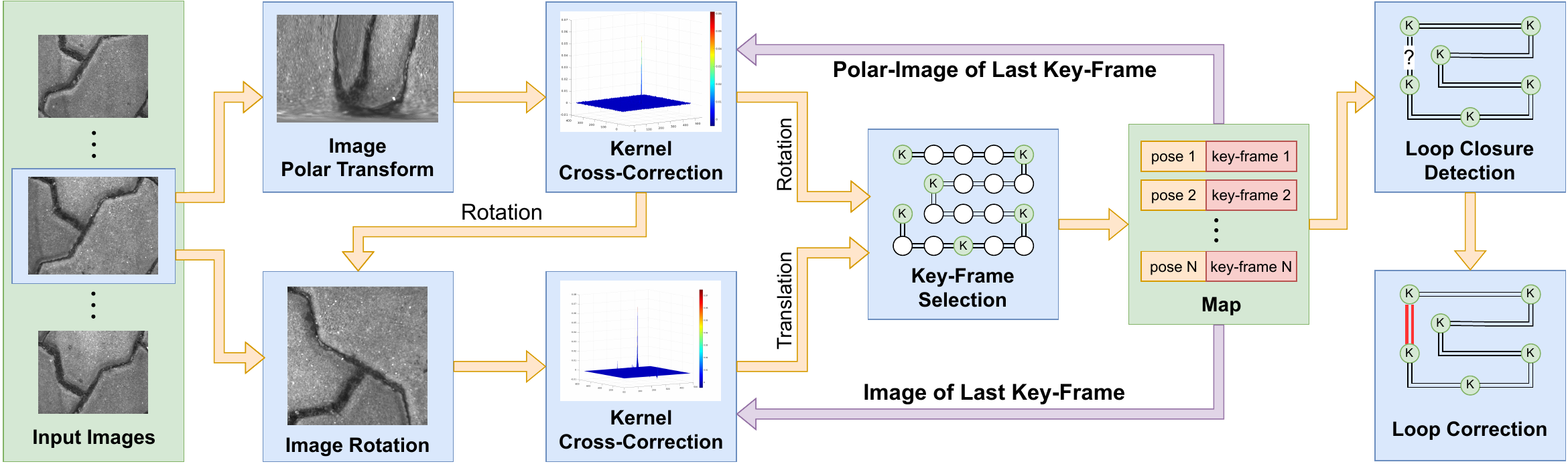}
    \caption{The pipeline of our GroundSLAM system. In the front-end, rotation and translation are decoupled and estimated using kernel cross-correlators. In the back-end, a keyframe-based map is maintained to facilitate loop closure detection and correction.}
    \label{fig:pipeline}
    \vspace{0.1em}
\end{figure*}

\subsubsection*{Estimation Confidence:}

Due to the boundary effect and noises, we cannot obtain the same correlation output defined in \eqref{eq:single-peak}, however, we are able to estimate its confidence from the significance of its peak. In the experiments, we use the peak-to-sidelobe ratio \citep{Henriques:2015jy} as a measure:
\begin{equation}\label{eq:confidence}
  P_s(\mathbf{x}) = \frac{\max \mathbf{g}(\mathbf{x},\mathbf{h}^\star)-\mu_s(\mathbf{g}(\mathbf{x},\mathbf{h}^\star))}{\delta_s(\mathbf{g}(\mathbf{x},\mathbf{h}^\star))},
\end{equation}
where $\mu_s$ and $\delta_s$ are the mean and standard deviation of the sidelobe, which is the output $\mathbf{g}(\mathbf{x},\mathbf{h}^\star)$ excluding the peak.

It is worth noting that KCC solutions \eqref{eq:kernel-solution} and \eqref{eq:kcc-solution} are the first generalized solutions that are capable of estimating any affine transformations including translation and rotation. 
It can also be used for weighted signals and any kernel functions, which are not researched in this paper.
All the other correlators such as MOSSE \citep{Anonymous:2010tm} and KCF \citep{Henriques:2015jy} can only estimate translations due to their theoretical limitations.
For simplicity, we present their limitations in \appref{app:appendix-kcf}, illustrating the theoretical significance of KCC. 
We next show how KCC can be applied to GroundSLAM with an efficient system architecture.

\section{System Architecture}\label{sec:system_architecture}

\subsection{System Overview} 

The overall structure of this framework is illustrated in \fref{fig:pipeline}. The system takes grayscale images captured by a downward-facing camera as input and estimates 3-DOF camera poses, consisting of 2-DOF translation and rotation.  
The workflow follows a structured process: The initial frame is designated as the first keyframe. For each new frame, the relative motion with respect to the previous keyframe is estimated. Based on the computed relative pose and its confidence, a new keyframe is selected and incorporated into the map. The system then searches for neighboring keyframes within the map to detect potential loop closures. If a loop closure is identified, pose graph optimization is performed to correct drift errors and improve localization accuracy.  
In the following sections, we provide a detailed explanation of each module within the framework.

\subsection{Efficient Kernel Vector} \label{sec:kernel_vector}

Building upon the result in \thmref{thm:transform}, which shows that any affine transformation can be estimated using KCC. We next present an efficient method for estimating the translation and rotation relative to the keyframews within the GroundSLAM framework.
Specifically, this requires computing the kernel vectors $\boldsymbol{\kappa}_{\mathbf{z}}(\mathbf{x})$ for translation and rotation transformations $\mathcal{T}_i$ in the frequency domain.
Without loss of generality, we consider the Gaussian kernel as a representative example, where the $i$-th element of the kernel vector is given by
\begin{equation}\label{eq:kernel-ele}
  \begin{aligned}
    \boldsymbol{\kappa}_{\mathbf{z}}(\mathbf{x})[i] &= e^{-\frac{1}{2}\|\mathbf{x}-\mathcal{T}_i(\mathbf{z})\|}\\
    &= e^{\mathbf{x}^T\mathcal{T}_i(\mathbf{z}) - \frac{1}{2}\|\mathbf{x}\|^2 - \frac{1}{2}\|\mathcal{T}_i(\mathbf{z})\|^2}.
  \end{aligned}
\end{equation}
Since computing $\|\mathcal{T}_i(\mathbf{z})\|^2$ has a complexity of $\mathcal{O}(n)$, constructing the entire kernel vector $\boldsymbol{\kappa}_{\mathbf{z}}(\mathbf{x})$ results in an overall complexity of $\mathcal{O}(n^2)$. In the following, we demonstrate how this computation can be efficiently accelerated in the frequency domain, reducing the complexity to $\mathcal{O}(n\log n)$.

\paragraph{Translation} \label{sec:ktc}

Assume $\mathcal{T}_i(\mathbf{z})$ is a circular translation, i.e., $\mathcal{T}_i(\mathbf{z}) = \mathbf{z}_{(i)}$, then the kernel vector is $\boldsymbol{\kappa}_{\mathbf{z}}(\mathbf{x}) = [\kappa(\mathbf{x}, \mathbf{z}_{(0)}), \cdots, \kappa(\mathbf{x}, \mathbf{z}_{(m-1)})]^T$. For each element in the kernel vector, we can take out the common items, then
\begin{equation}
  \begin{aligned}
    \boldsymbol{\kappa}_{\mathbf{z}}(\mathbf{x}) &= e^{-\frac{1}{2}\|\mathbf{x}\|^2 - \frac{1}{2}\|\mathbf{z}\|^2} \cdot e^{{\big[ \mathbf{x}^T\mathbf{z}_{(0)},~\cdots,~\mathbf{x}^T\mathbf{z}_{(m-1)}\big]}^T},\\
    &= e^{-\frac{1}{2}\|\mathbf{x}\|^2 - \frac{1}{2}\|\mathbf{z}\|^2} \cdot e^{\mathbf{x}\otimes\phi(\mathbf{z})},
  \end{aligned}
\end{equation}
where $\|\mathbf{z}\|^2  = \|\mathbf{z}_{(i)}\|^2$ due to the circular translation property. Therefore, the kernel vector includes another cross-correlation which can be computed in the frequency domain
\begin{equation}
  \begin{aligned}
    \boldsymbol{\kappa}_{\mathbf{z}}(\mathbf{x})
    &= e^{-\frac{1}{2}\|\mathbf{x}\|^2 - \frac{1}{2}\|\mathbf{z}\|^2} \cdot e^{\mathcal{F}^{-1}\big(\hat{\mathbf{x}}\odot\hat{\mathbf{z}}\big)},
  \end{aligned}
\end{equation}
of which the complexity is reduced to $\mathcal{O}(n\log n)$ from $\mathcal{O}(n^2)$. The sum of squares can also be computed in the frequency domain using Parseval's theorem \citep{parseval1806memoire}:
\begin{equation}\label{eq:kernel-vector-fft}
  \begin{aligned}
    \boldsymbol{\kappa}_{\mathbf{z}}(\mathbf{x})
    &= e^{-\frac{1}{2n}\big(\|\hat{\mathbf{x}}\|^2 - \|\hat{\mathbf{z}}\|^2\big)} \cdot e^{\mathcal{F}^{-1}\big(\hat{\mathbf{x}}\odot\hat{\mathbf{z}}\big)}.
  \end{aligned}
\end{equation}
Equation \eqref{eq:kernel-vector-fft} is able to further save the computer memory when only the FFT of the keyframes are retained.

\paragraph{Rotation} \label{sec:rotation_polar}

Different from translation, we cannot convert the kernel vector to the frequency domain directly.
However, KCC doesn't specify the transformation function $\mathcal{T}_i$, we can design the kernel vector leveraging the periodicity of rotation.
Assume $\mathbf{x}$ and $\mathbf{z}$ are 2D images and denote the Cartesian coordinate as $[x,y]$ and polar coordinate as $<\rho,\theta>$. If we define $\mathcal{T}_i(\mathbf{z})[j,k]=\mathbf{z}<\rho_j, \frac{2i\pi}{m}+\frac{2k\pi}{N}>$, where $i\in 0,\cdots, m-1$, $j\in 0,\cdots, M-1$, $k\in 0,\cdots, N-1$ and $\rho_j \in [0, \frac{\min(M,N)}{2}]$, then columns of $\mathcal{T}_i(\mathbf{z})$ are circular vectors and the rotation kernel vector become
\begin{equation}\label{eq:rotation-vector}
  \begin{aligned}
    \boldsymbol{\kappa}_{\mathbf{z}}(\mathbf{x})
    &= e^{-\frac{1}{2}\|\mathbf{x}\|^2 - \frac{1}{2}\|\mathcal{T}_0(\mathbf{z})\|^2} \cdot e^{\sum_{k=0}^{N-1}(\mathbf{x}[:,k]\otimes\mathcal{T}_0(\mathbf{z})[:,k])}\\
    &= e^{-\frac{1}{2n}\|\hat{\mathbf{x}}\|^2 - \frac{1}{2n}\|\hat{\mathcal{T}}_0(\mathbf{z})\|^2} \cdot e^{\sum_{k=0}^{N-1}\mathcal{F}^{-1}\big(\hat{\mathbf{x}}[:,k]\odot\hat{\mathcal{T}}_0(\mathbf{z})[:,k]\big)}
  \end{aligned},
\end{equation}
which can also be computed in the frequency domain ($\mathbb{O}(n\log n)$) independently and each column of $\mathcal{T}_i(\mathbf{z})[:,k]$ is a circular vector. Additionally, we take the summation of its columns, thus the kernel vector $\boldsymbol{\kappa}_{\mathbf{z}}(\mathbf{x})\in\mathbb{R}^m$ is a 1D signal, further reducing the computational complexity ($m\ll n$).
% , regardless that the images $\mathbf{x},\mathbf{z}$ are matrices.

\begin{figure}[t]
    \vspace{0.4em}
    \centering
    \includegraphics[width= 0.99\linewidth]{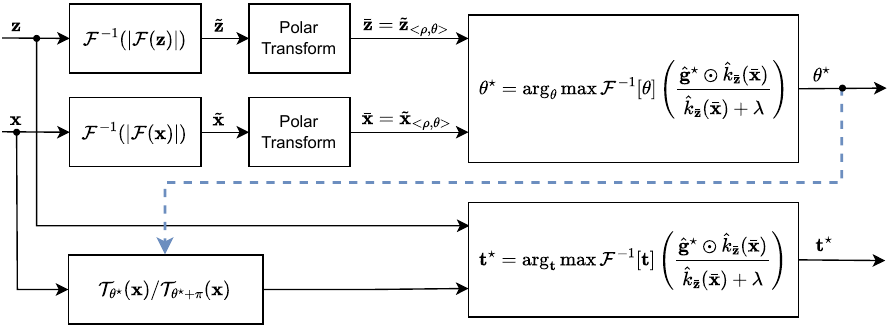}
    \caption{An illustration of relative transformation estimation using the proposed KCC. The keyframe and the current image are represented as $\mathbf{z}$ and $\mathbf{x}$, respectively. Their relative rotation $\mathbf{\theta}$ and translation $\mathbf{t}$ are estimated in a decoupled way.}
    \label{fig:motion-estimation}
    % \vspace{-0.9em}
\end{figure}

\subsection{Motion Estimation} \label{sec:motion_estimation}

Equation \eqref{eq:kernel-vector-fft} and \eqref{eq:rotation-vector} are able to estimate the translation and rotation movement independently, but they often couple with each other in the real-world. We next show how translation can be decoupled from rotation so that \eqref{eq:kernel-vector-fft} and \eqref{eq:rotation-vector} can be applied sequentially, as shown in \fref{fig:motion-estimation}.

\paragraph{Movement Decoupling} \label{sec:decoupling}

% As shown in \fref{fig:motion-estimation}, to apply KCC in an efficient way, we decouple the rotation and translation movement, which is 
Inspired by the properties of FFT: translation in the spatial domain corresponds to phase shift in the frequency domain, while rotation in the spatial domain corresponds to the same rotation in the frequency domain, we can eliminate the effect of translations by taking the FFT magnitude \citep{tong2019image}.
Suppose $\mathbf{x}$ is a translation and rotation of $\mathbf{z}$, so we have $\mathbf{x}[i, j] = \mathbf{z}[i\cos\theta^\star+j\sin\theta^\star-i^\star, j\cos\theta^\star-i\sin\theta^\star-j^\star]$, where $\mathbf{t}^\star = (i^\star, j^\star)$ is the translation and $\theta^\star$ is the rotation angle, their FFT is $\hat{\mathbf{x}} [\zeta, \eta] = \hat{\mathbf{z}}[\zeta\cos\theta^\star+\eta\sin\theta^\star, \eta\cos\theta^\star-\zeta\sin\theta^\star]e^{-2\pi(\zeta i^\star, \eta j^\star)}$.  Therefore, the translation can be decoupled from rotation by only taking the FFT magnitude, i.e., $|\hat{\mathbf{x}} (\zeta, \eta)| = |\hat{\mathbf{x}}[\zeta\cos\theta^\star+\eta\sin\theta^\star, \eta\cos\theta^\star-\zeta\sin\theta^\star]|$. 
In other words, there is only a rotation between $\tilde{\mathbf{x}}$ and $\tilde{\mathbf{z}}$:
\begin{equation} \label{eq:new_image}
    \tilde{\mathbf{x}} = \mathcal{F}^{-1}(|\hat{\mathbf{x}}|),\quad \tilde{\mathbf{z}} = \mathcal{F}^{-1}(|\hat{\mathbf{z}}|),
\end{equation}
where $\mathcal{F}^{-1}$ is the inverse FFT (IFFT) function.
In this context, rotation and translation can be treated as independent components, with rotation being estimated first.
To achieve this, each newly captured image is processed to generate a transformed version using \eqref{eq:new_image}. The transformed image is then remapped to the polar domain, referred to as the polar image. Leveraging \thmref{thm:transform}, the relative rotation $\theta^\star$ between the two transformed images is estimated. The confidence of the rotation estimation, quantified by the confidence score $P_s^r$, is subsequently determined using \eqref{eq:confidence}. Subsequently, the current frame can be rotated according to the estimated rotation $\theta^\star$ to compute the translation.

\paragraph{Ambiguity Resolution} 
Recall that if a signal in one domain is real, then the signal in the frequency domain has to be symmetric, thus $\tilde{\mathbf{x}},\tilde{\mathbf{z}}$ are symmetric images, which will pose challenges for rotation estimation.
Due to this ambiguity problem, the real rotation $\theta_I$ could be $\theta^\star$ or $\theta^\star+\pi$. In the tracking stage, we have a prior assumption that the rotation between two nearby frames is not significant, therefore, the one with the smaller absolute value between $\theta^\star$ and $\theta^\star+\pi$ is selected as $\theta_I$. Then we rotate the current image by $\theta_I$ and compute the transformation between it and the last keyframe using \thmref{thm:transform}. The estimated translation and the estimation confidence are denoted as $\mathbf{t}_I$ and $P_s^t$ respectively.

However, in the loop detection and map reuse stage, the prior assumption mentioned above is not valid. So to resolve the ambiguity, we rotate the current frame by $\theta^\star$ and $\theta^\star+\pi$, computing two translations $\mathbf{t}_1^\star$ and $\mathbf{t}_2^\star$ and their corresponding confidences $P_s^{t_1}$ and $P_s^{t_2}$ using \thmref{thm:transform} and \eqref{eq:confidence}, respectively. The final transform can be determined by 
\begin{equation}
    \left( \theta_I, \mathbf{t}_I, P_s^t \right) = \left\{ \begin{array}{lcl}
        \left( \theta^\star, \mathbf{t}_1^\star, P_s^{t_1} \right) & \mbox{for} & \text{$P_s^{t_1} >= P_s^{t_2}$} \\
        \left( \theta^\star+\pi, \mathbf{t}_2^\star, P_s^{t_2} \right) & \mbox{for} & \text{$P_s^{t_2} > P_s^{t_1}$}.
                                          \end{array}\right.
\end{equation}

\paragraph{Camera Pose}
Note that $\theta_I$ and $\mathbf{t}_I$ are the rotation angle and 2D translation on the image plane, with the coordinate origin being at the image center. To get the relative camera pose, we first compute the rotation $\theta_o$ and the translation $\mathbf{t}_o$ with the principal point being the origin:
\begin{subequations}\label{eq:rt_principal}
    \begin{align}
        \theta_o &= \theta_I ,\\
        \mathbf{t}_o &= \left(\mathbf{I}_{2\times2}-\mathbf{R}(\theta_I) \right) 
                \renewcommand{\arraystretch}{1.2}
                \left[\begin{array}{c} 
                    W/2-c_x \\ H/2-c_y 
                \end{array}\right] + \mathbf{t_I},
    \end{align}
\end{subequations}
where $W$, $H$, and $\left(c_x, c_y\right)$ are the width, height, and principal point of the image, respectively. $\mathbf{I}_{2\times2}$ denotes the 2D identity matrix. $\mathbf{R}(\theta_I)$ is the rotation matrix generated by 
\begin{equation}
    \mathbf{R}(\theta_I) = \left[\begin{array}{cc} \cos{\theta_I} & -\sin{\theta_I} \\ \sin{\theta_I} & \cos{\theta_I} \end{array}\right].
\end{equation}
Then the relative camera pose can be obtained by
\begin{equation}\label{eq:rt_camera}
    \left[\begin{array}{ccc} \theta_c &  t_{c_x} &  t_{c_y} \end{array}\right]^{\mathrm{T}} =
        \left[\begin{array}{ccc} \theta_o &  \frac{h_c}{f_x} t_{o_x} &  \frac{h_c}{f_y} t_{o_y} \end{array}\right]^{\mathrm{T}},
\end{equation}
where $\theta_c$ and $\mathbf{t_c}$ are the camera rotation (the yaw angle) and translation, respectively, $f_x$ and $f_y$ are the focal lengths and $h_c$ is the height of the camera from the ground, which can be obtained from the camera extrinsic calibration.

\subsection{Keyframe Selection}
The motion estimations described in \sref{sec:motion_estimation} are determined at the pixel level. Therefore, to reduce drift errors, keyframes are strategically selected, and relative pose estimation is performed only between the current frame and the most recent keyframe. A frame is designated as a keyframe if any of the following conditions are satisfied:
\begin{itemize}[noitemsep,topsep=0pt]
    \item The distance to the last keyframe in the normalized image plane is larger than $\delta_d$;
    \item The angle with the last keyframe is larger than $\delta_\theta$;
    \item The confidence of rotation estimation falls within a predetermined range, \ie $P_s^{r_{min}} < P_s^r < P_s^{r_{max}}$; and
    \item The confidence of translation estimation falls within a predetermined range, \ie $P_s^{t_{min}} < P_s^t < P_s^{t_{max}}$,
\end{itemize}
where $\delta_d$, $\delta_\theta$, $P_s^{r_{min}}$, $P_s^{r_{max}}$, $P_s^{t_{min}}$ and $P_s^{t_{max}}$ are all preset thresholds.
The first two conditions ensure sufficient overlap between the two frames, while the last two conditions balance the trade-off between reducing the number of keyframes and preserving the reliability of the estimation.

\subsection{Loop Detection}
To correct the drift error, the loop closure should be detected if the robot goes back to a previously visited place. We maintain an online map that uses a hash table to store the indices, poses, FFT results, and polar images of keyframes. Once a new keyframe is inserted into the map, two kinds of verification are utilized to find a real loop.

\paragraph{Geometric Verification}
We first retrieve all the keyframes within a distance of $d_l$ of the current keyframe as loop candidates. To avoid the false loop pair with neighboring keyframes, candidates with a travel distance less than $\delta_{d}^l$ or an index difference less than $\delta_N^l$ from the current frame will be removed. Generally speaking, geometric verification can rapidly filter out the majority of outliers.

\paragraph{Correction Verification}
For each remaining keyframe, the relative pose to the current keyframe will be computed. The candidate with the highest confidence is taken as the best loop candidate, and if its rotation estimation confidence is higher than $P_s^{r_{\mathrm{min}}}$ and translation estimation confidence is higher than $P_s^{t_{\mathrm{min}}}$, it will be considered as a valid loop closure.

Upon detecting a valid loop closure, a pose graph consisting of keyframes is constructed. This graph comprises two types of edges: the odometry edges, which connect consecutive keyframes, and the loop closure edges, which connect keyframes that form a loop closure. The graph is then optimized using the Levenberg-Marquardt algorithm \citep{levenberg1944method} to refine the estimated poses. Once the optimization is completed, the poses of keyframes in the map are updated to improve overall trajectory consistency.

\begin{algorithm}[t]
    \small 
    \renewcommand{\algorithmicrequire}{\textbf{Input:}}
    \renewcommand{\algorithmicensure}{\textbf{Output:}}
    \caption{Map Reuse}
    \label{alg:map_reuse}
    \begin{algorithmic}[1]
        \REQUIRE
            $\mathbf{x}$: the current image; 
            $\left\{\mathbf{x}_i^k, i < N \right\}$: the candidates;
            $\left(  P_s^{r_{min}},  P_s^{t_{min}} \right)$: confidence thresholds;
        \ENSURE
            $\left(\theta,  \mathbf{t} \right)$: the current pose, $P_s$: the confidence; 
        \STATE Initialization: $i=0$ and $P_s=0$;
        \FOR {$i < N$}
            \STATE $i \leftarrow i + 1$
            \STATE $\Delta{\theta_i}, P_s^{r_i} = \mathrm{RotationEstimation}(\mathbf{x}_i^k, \mathbf{x})$
            \IF {$P_s^{r_i} > P_s^{r_{min}}$}
                \STATE  $\Delta\mathbf{t}_i, P_s^{t_i} = \mathrm{TranslationEstimation}(\mathbf{x}_i^k, \mathbf{x})$
                \IF {$P_s^{t_i} > P_s^{t_{min}}$ and $P_s < (P_s^{r_i} + P_s^{t_i})$}
                    \STATE $P_s \leftarrow P_s^{r_i} + P_s^{t_i}$
                    \STATE $\theta \leftarrow \theta_i^k + \Delta{\theta_i}$
                    \STATE $\mathbf{t} \leftarrow \mathbf{t}_i^k + \mathbf{R}(\theta_i^k) \Delta\mathbf{t}_i$
                \ENDIF
            \ENDIF
        \ENDFOR
    \end{algorithmic}
\end{algorithm}

\subsection{Map Reuse} \label{sec:map_reuse}

In practical applications such as warehouse logistics, a prior map is typically constructed in advance, allowing multiple robots to perform drift-free localization using the same map. In this section, we present an approach for storing and reusing a prior map. Unlike existing methods that rely on various feature descriptors and Bag of Words \citep{galvez2012bags}, our map reuse module leverages KCC to compute frame-to-frame similarity. Consequently, unlike conventional visual maps that require storing point clouds, feature descriptors, and observation relationships, the proposed system maintains a more lightweight and structured representation. Specifically, we store only the keyframe poses, their FFT results, and polar-transformed images, significantly reducing memory requirements while preserving localization accuracy. Notably, the keyframe poses can be obtained not only from the proposed system but also from alternative localization sources, such as other visual or LiDAR SLAM systems \citep{wang2021f} and ultra-wideband (UWB) positioning \citep{wang2017ultra}.

When deploying the constructed map in the warehouse, we adopt a local localization strategy rather than global localization to enhance both accuracy and efficiency. For each query image, we use a prior pose to retrieve all neighboring keyframes within a distance of $d_{mr}$ as candidates. 
The threshold $d_{mr}$ is determined based on the covariance of the prior pose. Subsequently, we apply \algref{alg:map_reuse} to compute the current pose and the associated confidence $P_s$. The estimation is considered valid if $P_s > 0$.

\begin{figure}[t]
    \vspace{-0.1em}
    \centering
    \includegraphics[width=0.99\linewidth]{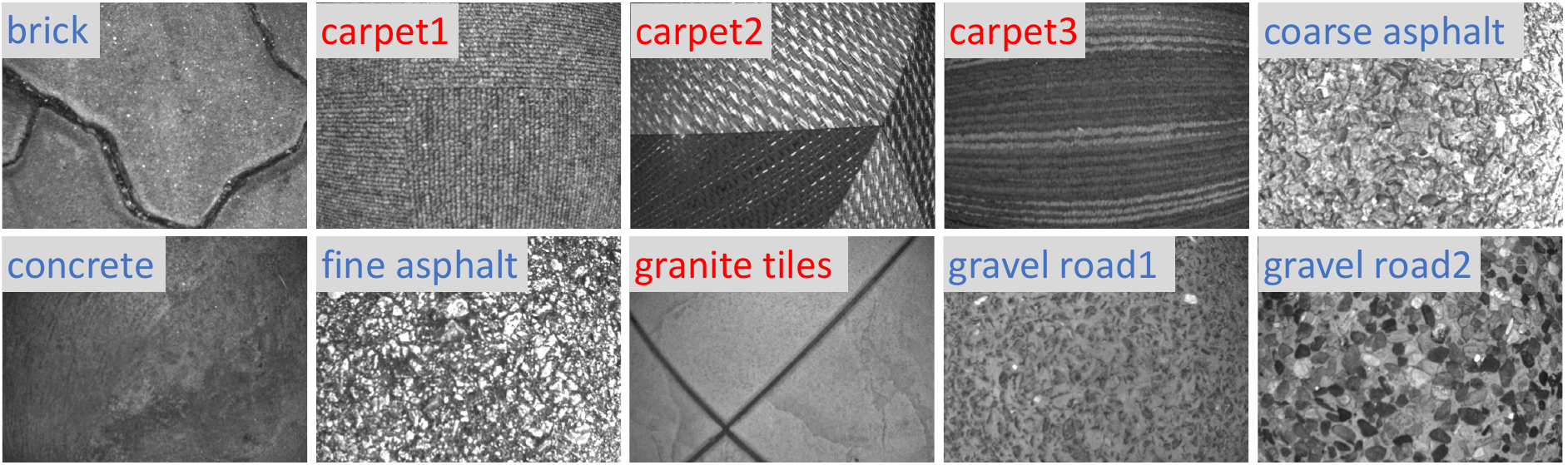}
    \caption{Example ground texture images in our dataset. We collect 10 kinds of ground textures, including 6 kinds of {\color{NavyBlue}{outdoor}} textures and 4 kinds of \textcolor{red}{indoor} textures.}
    \label{fig:sample_images}
    \vspace{-0.75em}
\end{figure}

\begin{table}[!t]
    \caption{A comparison between our PathTex dataset and the HD Ground dataset \citep{schmid2022hd}.}
    % \vspace{-5pt}
    \label{tab:public_dataset}
    \centering
    \resizebox{\linewidth}{!}{
    \begin{tabular}{ccc}
      \toprule
       Datasets                     & HD Ground \citep{schmid2022hd}  & PathTex (Ours)     \\
      \midrule
      Images                        & 201428      & 131300  \\
      Images Per Meter              & 14.98       & 242.28  \\
      Avg Overlap Rate              & 47.74\%     & 93.69\% \\
      Textures Types                & 11          & 10      \\
      % Textures with Long Sequence   & 7           & 10      \\
      % Avg Sequence Length  (\meter) & 11.18       & 30.11   \\
      % Max Sequence Length (\meter)  & 55.09       & 55.47   \\
      Ground Truth                  & Image Stitching & Leica Total Station \\
      \bottomrule
    \end{tabular}
    }
    \vspace{0.1em}
\end{table}

\section{Experiments} \label{sec:experiments}

The remainder of this section is organized as follows. In \sref{sec:datasets_and_baselines_e}, we introduce the baselines and datasets used in the following experiments. Our field-collected dataset, the PathTex dataset, is also introduced.
We then evaluate the performance on data association, visual odometry, loop detection, map reuse, and real-world demo in \sref{sec:data_association_e}, \ref{sec:vo_e}, \ref{sec:loop_e}, \ref{sec:map_reuse_e}, and \ref{sec:live-demo}, respectively.
The ablation study on the effects of the kernel method and analysis on system efficiency are presented in \sref{sec:ablation_study_e} and \ref{sec:efficiency_analysis_e}, respectively.

\begin{figure*}[t]
    \vspace{0.5em}
    \centering
    \setlength{\abovecaptionskip}{0.4em}
    \includegraphics[width=0.989\linewidth]{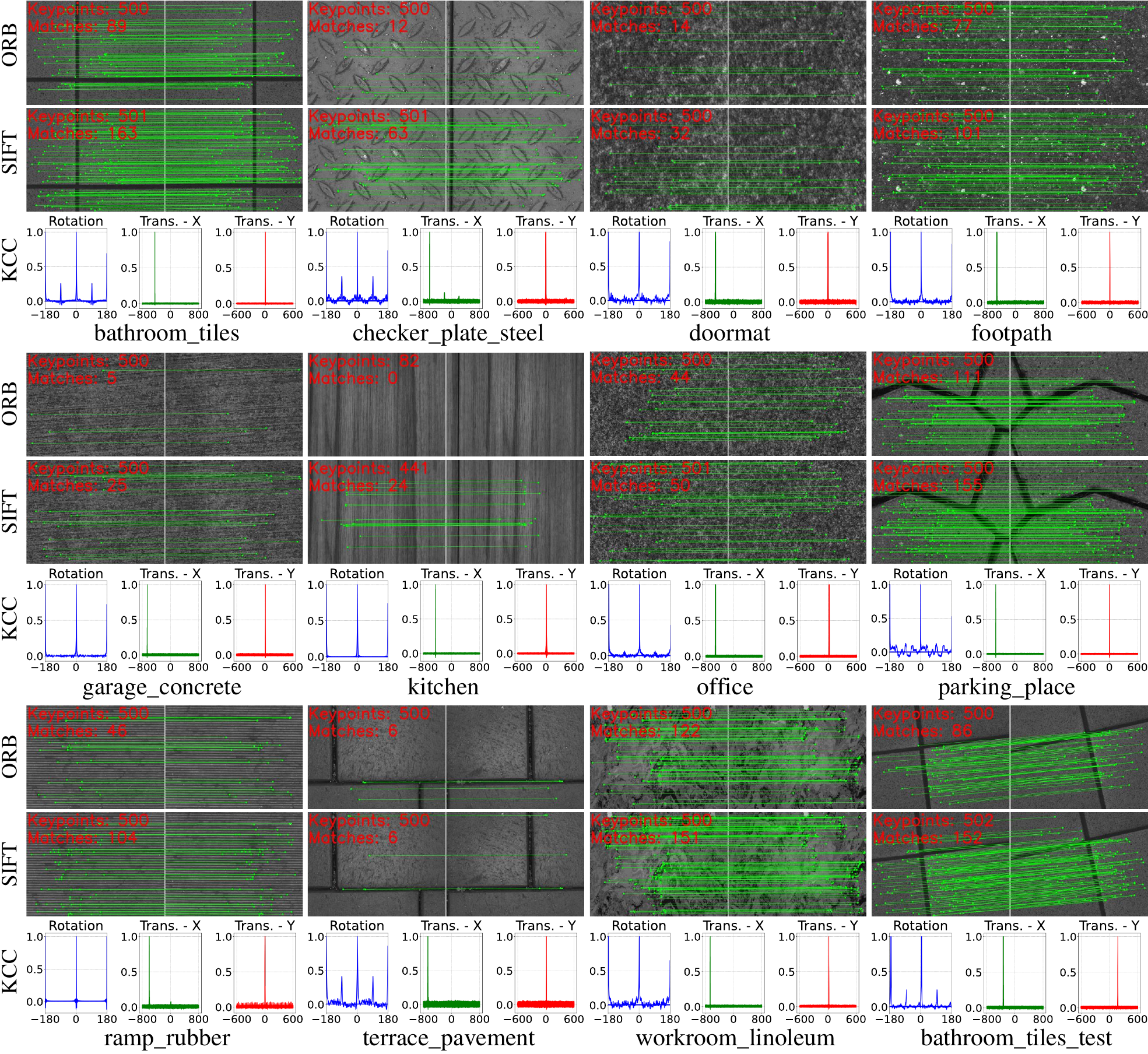}
    \caption{The comparison of data association of ORB, SIFT, and KCC on the HD Ground dataset. The numbers of features and matching inliers are given. For the KCC, the correction results are projected to three coordinate axes and represent the estimation of the 3-DOF movement. The vertical axis is the confidence of estimated movement on the horizontal axis.  The higher the value of the peak relative to other positions, the greater the confidence of motion estimation.}
    \label{fig:feature_tracking}
    \vspace{-0.8em}
\end{figure*}

\subsection{New Datasets and Baselines} \label{sec:datasets_and_baselines_e}

\paragraph{Datasets} 
To our current understanding, there exist only two publicly accessible ground texture datasets: Micro-GPS \citep{zhang2019high} and the HD Ground dataset \citep{schmid2022hd}. Notably, both these datasets primarily target the evaluation of relocalization systems that utilize prior maps. This design inclination yields two specific attributes that render them \textit{less effective} for the assessment of VO and SLAM systems:
(1) Limited Overlap: These datasets aim for wide area coverage with fewer images, emphasizing visual aliasing effects \citep{schmid2022hd}. In pursuit of this objective, they re-sample the original image sequences, producing new sequences. However, the sampling strategy leads to minimal overlap between successive frames, making them not ideal for evaluating VO and SLAM systems.
(2) Ground Truth Accuracy Concerns: The datasets rely on image stitching to establish their ground truths, as opposed to leveraging precise pose measurement instruments. This means that their accuracy is largely dependent on the feature detection and matching in the image-stitching algorithm. 
However, we found that our KCC-based image matching can even produce higher accuracy than the feature-based matching (will be detailed in \sref{sec:data_association_e}).
This suggests that those datasets might not be sufficiently robust for evaluating accuracy.
Consequently, they are only used to evaluate data association, map reuse, and the success tracking rate. 
In addition, since the HD Ground dataset contains significantly more comprehensive data than the Micro-GPS dataset, we use only the HD Ground dataset for the evaluation.

To facilitate a comprehensive evaluation of VO and SLAM systems, we introduce the PathTex dataset. We utilized a modified Weston SCOUT Robot as our primary data collection platform. This robot is outfitted with an IDS uEye monocular camera, strategically positioned at its base facing downward, maintaining a height of $0.1\meter$ above the terrain. To achieve consistent illumination, an array of LED lights encircles the camera, following the configuration of most warehouse robots.
For accurate ground truth measurements, we affixed a prism atop the robot, which is continuously monitored by a Leica Nova MS60 MultiStation laser tracker.
As depicted in \fref{fig:sample_images}, our PathTex dataset encompasses 10 prevalent ground textures, segmented into 6 outdoor and 4 indoor textures. A comparative analysis against the HD Ground dataset is presented in \tref{tab:public_dataset}. Herein, the overlap rate between successive frames is defined by their intersection over union (IOU).
A notable observation is that the average overlap rate between adjoining frames in our dataset is about 2 times greater than the HD Ground dataset, making our dataset ideal for evaluating the overall performance of the frame-by-frame VO and SLAM systems.

\begin{table*}[t]
    % \vspace{0.5em}
    \setlength\tabcolsep{3.5pt}
    \small
    \caption{\centering Translational error (RMSE) without loop closing on the PathTex dataset (unit: m). $\times$ refers to tracking lost. }
    \label{tab:ntu_ground_rmse}
    \centering  
    \begin{tabular}{C{0.19\linewidth}|C{0.06\linewidth}C{0.06\linewidth}C{0.1\linewidth}C{0.055\linewidth}C{0.06\linewidth}C{0.06\linewidth}C{0.08\linewidth}C{0.06\linewidth}C{0.06\linewidth}C{0.05\linewidth}}
        \toprule
         \multirow{2}{*}{Sequence} &  SURF- & SIFT- & \multirow{2}{*}{TartanVO} & \multirow{2}{*}{SVO} & ORB- &  \multirow{2}{*}{o-eFMT} & DROID- & LSD- & GT-  & \multirow{2}{*}{Ours}  \\ 
        & VO & VO & & & SLAM3 &  & SLAM & SLAM & SLAM  \\
        \midrule
        Brick\_seq1         & 0.634 & 0.545 & 1.043 & \underline{0.150} & \textbf{0.126} & 2.078 & 1.490 & 0.477 & 0.152 & 0.162 \\ 
        Brick\_seq2         & 0.503 & 0.480 & 1.223 & 0.337 & \underline{0.227} & 3.381 & 0.519 & 0.361 & 1.216 & \textbf{0.213} \\ 
        Carpet1\_seq1      & 3.987 & \underline{3.107} & 6.300 & 6.379 & $\times$ & 6.967 & 5.900 & 6.827 & 5.915 & \textbf{1.107} \\ 
        Carpet1\_seq2       & 2.248 & 1.825 & 5.013 & 1.152 & \textbf{0.145} & 5.371 & 4.690 & $\times$ & \underline{0.377} & 0.423 \\ 
        Carpet2\_seq1       & 1.716 & 1.482 & 2.890 & $\times$ & $\times$ & 3.265 & 2.898 & 2.063 & \underline{0.703} & \textbf{0.289} \\ 
        Carpet3\_seq1       & 0.348 & 0.349 & 1.178 & \underline{0.265} & $\times$ & 2.413 & 1.580 & 2.375 & 1.673 & \textbf{0.174} \\ 
        Carpet3\_seq2       & 0.637 & 0.597 & 1.052 & \underline{0.449} & $\times$ & 2.797 & 1.965 & 0.986 & 0.736 & \textbf{0.252} \\ 
        Carpet3\_seq3       & \underline{0.864} & 3.285 & 6.474 & $\times$ & $\times$ & 6.967 & 5.933 & 2.003 & 5.228 & \textbf{0.376} \\ 
        Coarse\_asphalt\_seq1 & 0.691 & 0.746 & 1.768 & \textbf{0.082} & \underline{0.120} & 2.199 & 0.705 & $\times$ & 0.217 & 0.163 \\ 
        Concrete\_seq1      & $\times$ & 1.010 & 1.680 & 0.776 & $\times$ & 3.771 & \underline{0.512} & 2.379 & 2.077 & \textbf{0.361} \\ 
        Concrete\_seq2      & $\times$ & $\times$ & 2.163 & 0.714 & $\times$ & 3.813 & \underline{0.565} & 2.608 & 0.664 & \textbf{0.262} \\ 
        Fine\_asphalt\_seq1 & 1.194 & 1.111 & 2.413 & \underline{0.241} & 0.266 & 2.842 & 2.743 & $\times$ & 1.033 & \textbf{0.160} \\ 
        Fine\_asphalt\_seq2 & 2.383 & 2.264 & 2.717 & $\times$ & $\times$ & 2.591 & 2.754 & 2.727 & \underline{2.136} & \textbf{0.171} \\ 
        Granite\_tiles\_seq1 & $\times$ & $\times$ & 1.377 & \textbf{0.180} & $\times$ & 2.005 & \underline{0.241} & 1.898 & 0.604 & 0.469 \\ 
        Granite\_tiles\_seq2 & $\times$ & 0.687 & 1.752 & \textbf{0.287} & $\times$ & 2.163 & 0.781 & 1.911 & 0.886 & \underline{0.518} \\ 
        Gravel\_road1\_seq1 & 0.928 & 1.086 & 1.603 & \underline{0.167} & 0.218 & 2.883 & 0.208 & 1.761 & 0.689 & \textbf{0.166} \\ 
        Gravel\_road2\_seq1 & 1.565 & 1.491 & 2.206 & \underline{0.414} & 0.422 & 2.623 & 1.905 & $\times$ & 0.855 & \textbf{0.339} \\
        \bottomrule
    \end{tabular}
    \vspace{-0.8em}
\end{table*} 

\begin{figure}[t]
    \vspace{0.0em}
    \centering
    \includegraphics[width=0.99\linewidth]{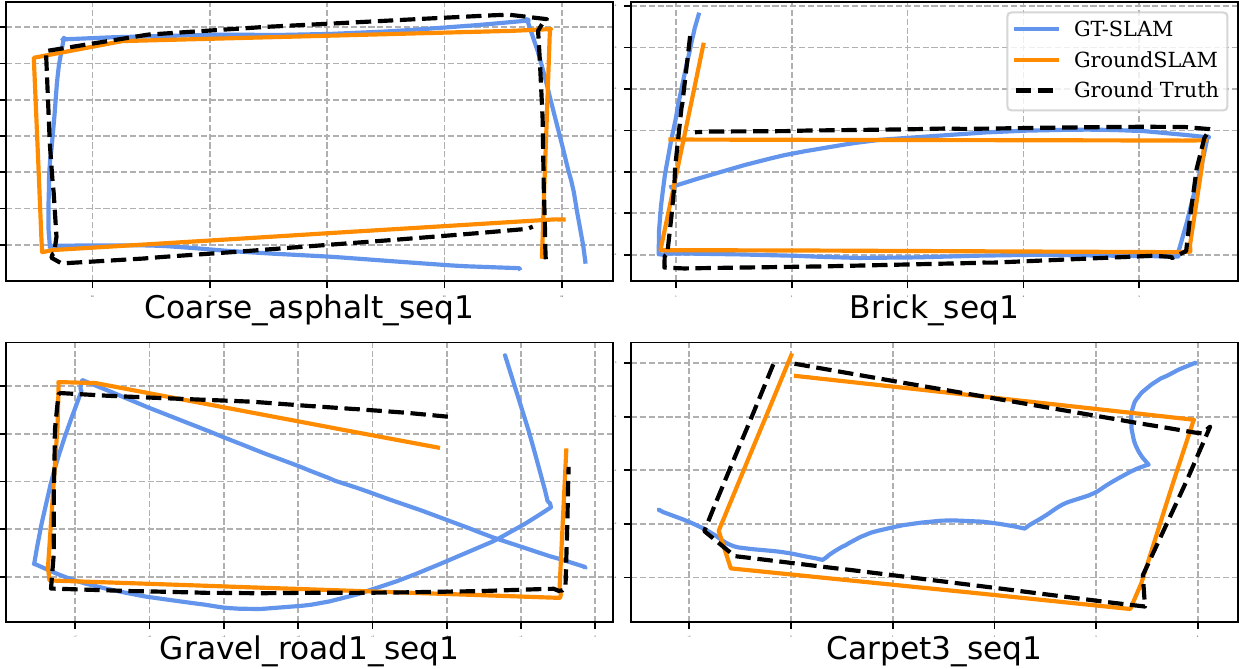}
    \caption{Estimated trajectories by our system (GroundSLAM) and GT-SLAM on four sequences of the PathTex dataset. The trajectories have been aligned with ground truth.}
    \label{fig:ntu-trajectory}
    \vspace{-1.5em}
\end{figure}

\paragraph{Baselines}

We evaluate our system against two categories of baseline methods:  
(1) Ground-texture-based systems. These methods estimate 2D rotation and translation for localization. We select two representative approaches: a correction-based method, o-eFMT \citep{jiang2023optimizing}, and a feature-based method, Ground-Texture-SLAM (GT-SLAM) \citep{hart2023monocular}. The o-eFMT system employs the Fourier-Mellin transform to estimate motion between consecutive images and integrates a back-end optimization module to enhance accuracy. GT-SLAM \citep{hart2023monocular}, on the other hand, utilizes ORB \citep{rublee2011orb} features for both front-end processing and loop closure detection, achieving a balance between efficiency and performance.  
Several other ground-texture-based localization systems \citep{chen2018streetmap, kozak2016ranger, schmid2020ground, nakashima2019akaze} are excluded from our comparison due to the unavailability of their source codes. To further investigate the impact of different feature extraction techniques, we implement visual odometry using SIFT \citep{lowe2004distinctive}, SURF \citep{bay2006surf}, and BRISK \citep{leutenegger2011brisk}, denoted as SIFT-VO, SURF-VO, and BRISK-VO, respectively.  
(2) Monocular visual SLAM systems. To assess the performance of widely used monocular SLAM frameworks, we compare our system against several established methods, including the feature-based ORB-SLAM3 \citep{campos2021orb}, the direct method SVO \citep{Forster17troSVO}, the learning-based image-level matching method TartanVO \citep{wang2021tartanvo}, the dense optical flow tracking-based DROID-SLAM \citep{teed2021droid}, and the photometric error optimization-based LSD-SLAM \citep{engel2014lsd}.  
It is important to note that, unlike ground-texture-based systems, these methods estimate full 6-DOF transformations rather than 3-DOF motion, potentially leading to increased computational overhead. While this difference makes direct comparisons less equitable, we nonetheless include these baselines for a broader evaluation.

\begin{figure}[t]
    \vspace{0.0em}
    \centering
    \includegraphics[width=0.99\linewidth]{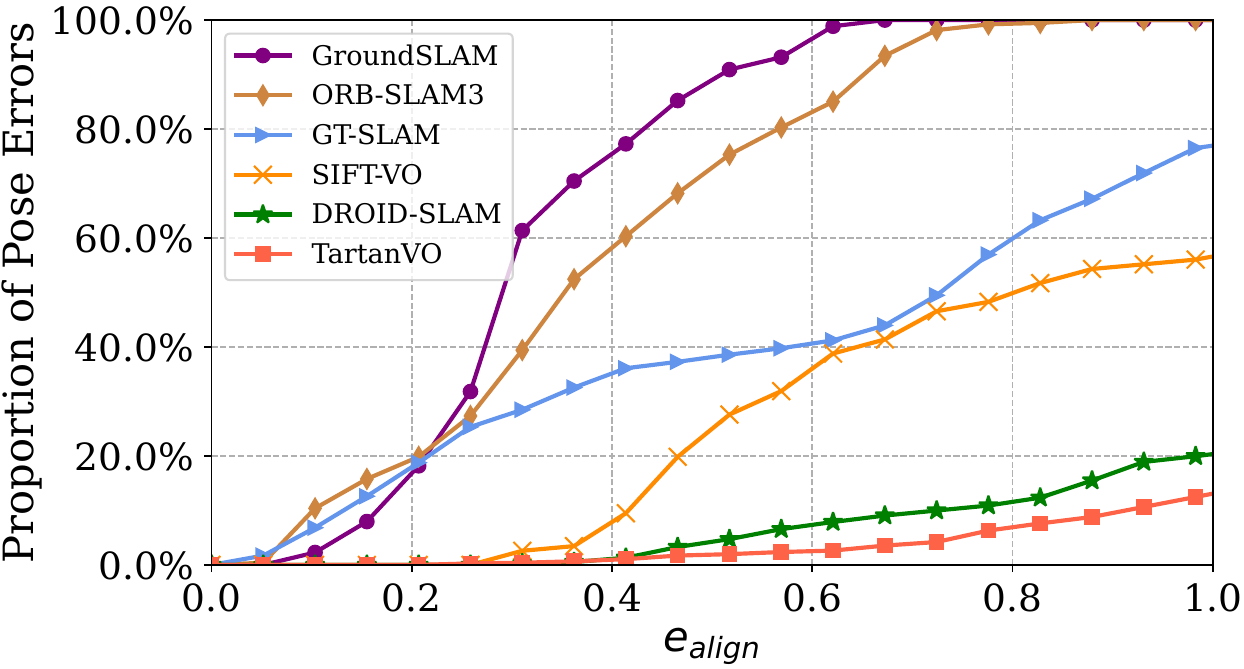}
    \caption{Comparison on ``Gravel\_road2\_seq1" from the PathTex dataset. The vertical axis shows the proportion of pose errors below the alignment error threshold on the horizontal axis.}
    \label{fig:ntu-rmse-curve}
    \vspace{-0.8em}
\end{figure}

\begin{table*}[t]
    % \vspace{0.5em}
    \setlength\tabcolsep{3.5pt}
    \small
    \caption{\centering Proportion of successful runs (\%) on the HD Ground dataset. The best results are \textbf{highlighted}.}
    \label{tab:hd_ground_rate}
    \centering  
    \begin{threeparttable}
    \begin{tabular}{C{0.12\linewidth}C{0.1\linewidth}|C{0.1\linewidth}C{0.08\linewidth}C{0.08\linewidth}C{0.08\linewidth}C{0.13\linewidth}C{0.08\linewidth}C{0.08\linewidth}}
        \toprule
        Texture & Sequences & BRISK-VO & SURF-VO & SIFT-VO & GT-SLAM & GT-SLAM$^{\text{w/o cvm}}$ & TartanVO & Ours  \\ 
        \midrule
        bathroom\_tiles & 31 & 38.7 & \textbf{100} & \textbf{100} & 61.3 & 61.3 & 0 & 96.8  \\ 
        checker\_plate\_steel & 12 & 8.3 & 25.0 & 41.7 & 0 & 0 & 0 & \textbf{75.0}  \\
        doormat & 25 & 40.0 & 64.0 & 64.0 & 4.0 & 4.0 & 0 & \textbf{76.0}  \\
        footpath & 18 & \textbf{100} & \textbf{100} & \textbf{100} & 88.9 & 88.9 & 88.9 & 94.4  \\ 
        garage\_concrete & 92 & 2.2 & 4.4 & 84.8 & 0 & 0 & 0 & \textbf{93.5}  \\ 
        kitchen & 25 & 0 & 24.0 & 24.0 & 64.0 & 4.0 & 84.0 & \textbf{100}  \\ 
        office & 24 & 87.5 & 75.0 & \textbf{100} & 83.3 & 79.2 & 83.3 & \textbf{100}  \\ 
        parking\_place & 25 & 96.0 & 88.0 & \textbf{100} & 84.0 & 68.0 & 72.0 & \textbf{100}  \\ 
        ramp\_rubber & 43 & 79.1 & 95.4 & \textbf{100} & 0 & 0 & 2.3 & \textbf{100}  \\ 
        terrace\_pavement & 49 & 0 & 0 & 0 & 0 & 0 & 0 & \textbf{71.4}  \\ 
        workroom\_linoleum & 60 & \textbf{100} & \textbf{100} & \textbf{100} & 30.0 & 30.0 & 0 & 76.7  \\
        \bottomrule
    \end{tabular}
        \begin{tablenotes}
            \footnotesize
            % \scriptsize
            \item[1] The results of SVO and ORB-SLAM3 are not listed as they fail to initialize on most sequences due to the small overlap between images.
        \end{tablenotes}
    \end{threeparttable}
    \vspace{-0.5em}
\end{table*}

\subsection{Data Association} \label{sec:data_association_e}

To benchmark the performance in low-overlap and different textures, we first evaluate the performance on data association using the HD Ground dataset.
The matching results of the ORB and SIFT features and our KCC are shown in \fref{fig:feature_tracking}.
Since KCC operates as an image-matching technique, its estimation is graphically represented with the normalized confidence on the vertical axis, while the horizontal axis details the corresponding movements in rotation and translation.
It can be seen that KCC consistently demonstrates accuracy and stability across diverse ground textures. This is evident from the distinctly pronounced peaks that tower significantly over other positions. 
However, the performance of feature-based methods fluctuates. They show commendable results on textures enriched with unique corners, such as in the sequence of ``bathroom\_tiles", ``footpath", ``parking\_place", and ``workroom\_linoleum"; while struggling to detect a sufficient number of matches in the sequence of ``checker\_plate\_steel", ``garage\_concrete", ``doormat", ``office", and ``terrace\_pavement". The situation further deteriorates with the ``kitchen" texture, where they fail to even detect an adequate number of features.

\begin{table}[t]
    \vspace{0.2em}
    \footnotesize
    \caption{Comparison of RMSE (unit: m) with and without loop correction on the PathTex dataset. }
    \label{tab:loop}
    \centering
    \begin{tabular}{C{0.24\linewidth}|C{0.12\linewidth}C{0.11\linewidth}|C{0.12\linewidth}C{0.11\linewidth}}
        \toprule
         \multirow{2}{*}{Sequence}  &  \multicolumn{2}{c|}{GT-SLAM}   & \multicolumn{2}{c}{GroundSLAM (Ours)}  \\ 
         & w/o loop & w/ loop  & w/o loop & w/ loop \\  \midrule
        Carpet1\_seq2       & 0.377 &  0.455 & 0.423 & \textbf{0.295}  \\ 
        Carpet2\_seq1       & 0.703 &  0.393 & 0.289 & \textbf{0.258}  \\
        Carpet3\_seq2       & 0.736 &  0.480 & 0.252 & \textbf{0.251}  \\
        Fine\_asphalt\_seq2 & 2.136 &  1.226 & 0.171 & \textbf{0.131} \\ 
        average             & 0.988 &  0.639 & 0.284 & \textbf{0.229}  \\ 
        \bottomrule     
    \end{tabular}
    \vspace{-0.8em}
\end{table}

\subsection{Visual Odometry} \label{sec:vo_e}

To compare the performance of frame-by-frame tracking and pose estimation among different SLAM systems, we conduct the odometry experiment where the loop detection and relocalization modules are removed from all the methods.

\paragraph{Accuracy on the PathTex Dataset} \label{sec:vo_ntu_e}

The efficacy of various systems, when benchmarked against the PathTex dataset, is evaluated using the Root Mean Square Error (RMSE) of translation and detailed in \tref{tab:ntu_ground_rmse}. Given that only positional ground truth data is accessible, a comparative analysis of rotational errors is omitted. The top-performing results are distinctly \textbf{highlighted} and \underline{underlined} in order. On an overarching scale, our system's resilience and precision stand unparalleled. It maintains consistent tracking across all sequences, securing minimal translational error in 12 out of the 17 sequences, indicating its best robustness over all other algorithms.
We notice that on some sequences such as ``Brick'', ``Coarse\_asphalt'', ``Granite\_tiles'', and ``Gravel\_road'', SVO and ORB-SLAM3 achieved slightly better performance than our system. 
This is because images of these sequences have more distinctive corners or edges, which are easy to track by feature-based methods. 
However, their dependence on prominent features or edges results in poor performance on the ground with minimal texture, leading to frequent tracking lost in scenarios like ``Concrete'', ``Carpet'', ``Fine\_asphalt'', and ``Granite\_tiles''.
In contrast, our GroundSLAM consistently delivers stable and dependable results \textit{across all sequences}.

\paragraph{Robustness Analysis}
We present the trajectories estimated by our system and GT-SLAM for four sequences in \fref{fig:ntu-trajectory}. The results show that our GroundSLAM exhibits significantly lower drift errors compared to GT-SLAM.  
A detailed comparison of error distributions for the ``Gravel\_road2\_seq1'' sequence is provided in \fref{fig:ntu-rmse-curve}. The vertical axis is the proportion of pose errors that are less than the given threshold on the horizontal axis. A higher proportion at a given threshold indicates better accuracy. The results reveal that the proportion of poses with errors below $0.3\meter$ is comparable among ORB-SLAM3, GT-SLAM, and our system. However, more than 90\% of the pose errors in our system remain below $0.5\meter$, demonstrating superior robustness and accuracy compared to the other methods.

\begin{figure}[t]
    \vspace{0.4em}
    \centering
    \setlength{\abovecaptionskip}{0.5em}
    \includegraphics[width=0.99\linewidth]{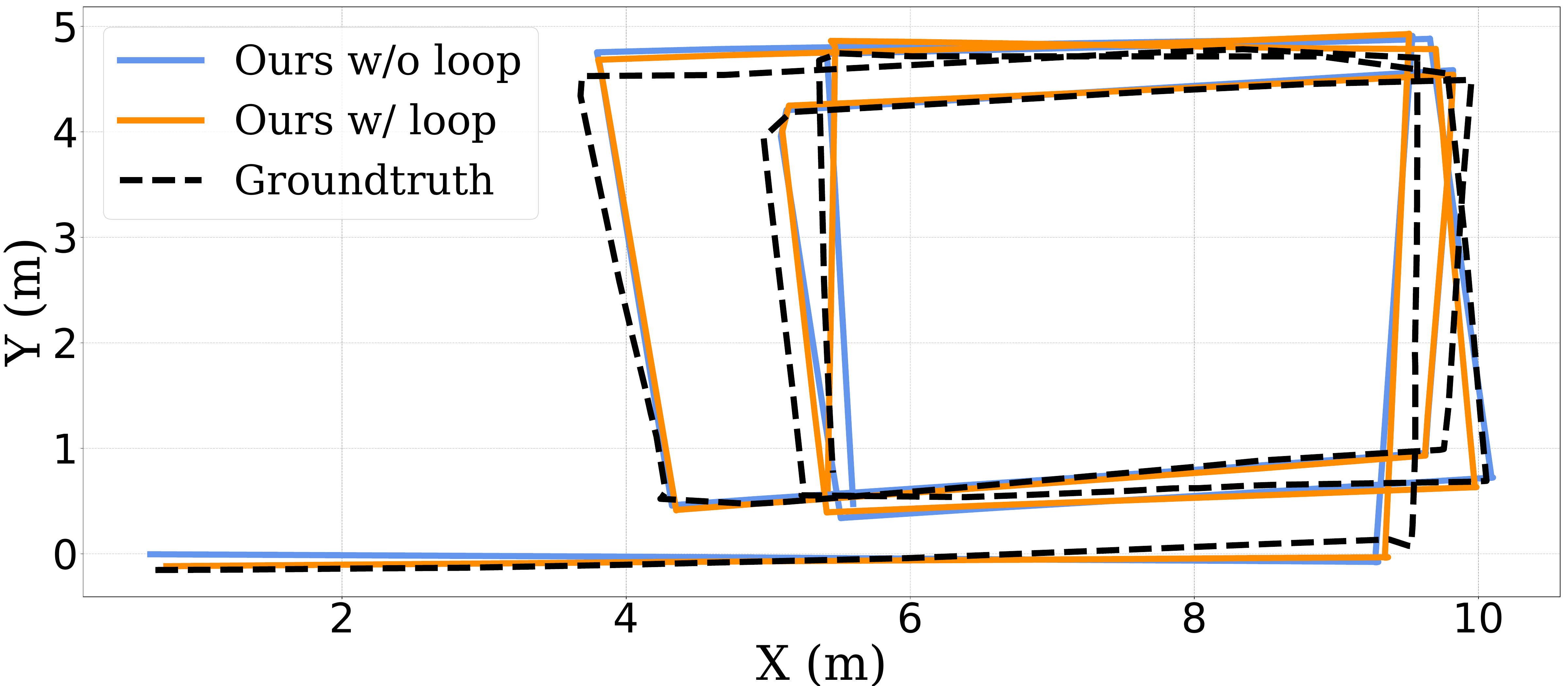}
    \caption{Trajectories estimated by our system with and without loop correction on the Fine\_asphalt\_seq2 sequence.}
    \label{fig:loop_trajectory}
    % \vspace{-1.1em}
\end{figure}

\begin{figure}[t]
    % \vspace{0.5em}
    \centering
    \setlength{\abovecaptionskip}{0.5em}
    \includegraphics[width=0.99\linewidth]{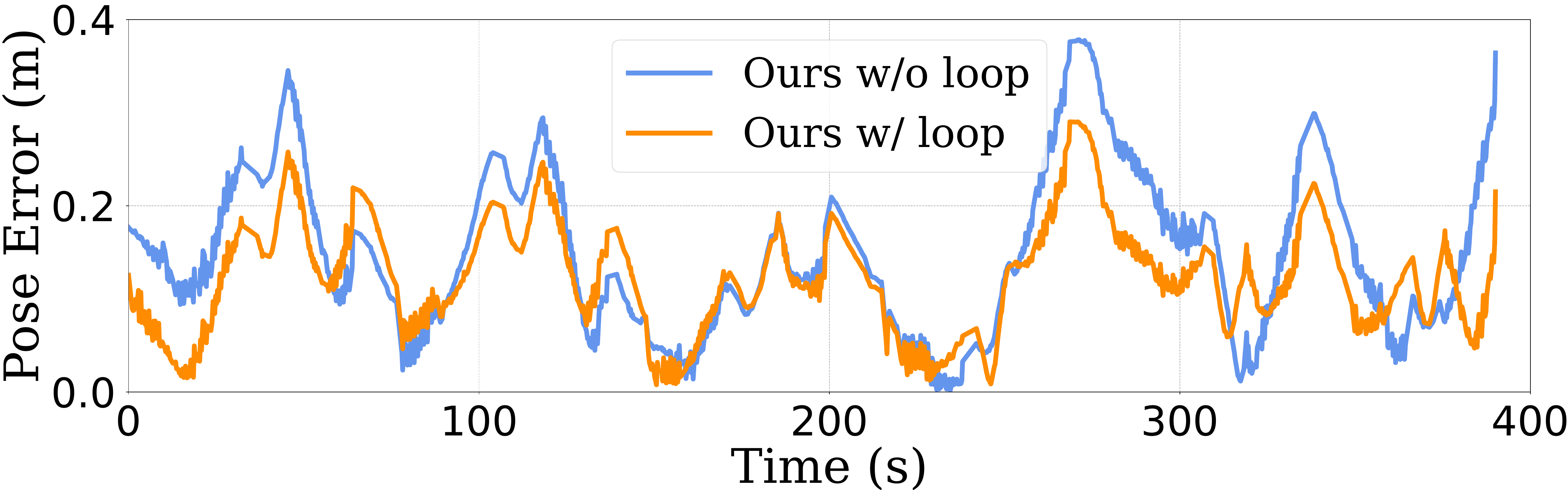}
    \caption{The pose errors (RMSE) of our system with and without loop correction on the Fine\_asphalt\_seq2 sequence.}
    \label{fig:loop_error}
    \vspace{-0.5em}
\end{figure}

Additionally, we find that although GT-SLAM and ORB-SLAM3 use the same ORB feature, there is a significant performance difference between these two systems. GT-SLAM is more robust while ORB-SLAM3 is more accurate. This may be due to their different feature tracking and optimization methods. ORB-SLAM3 predicts the camera pose with a constant velocity model and then projects 3D points onto the image to match the features. This strategy is more efficient but less robust than the FLANN-based method \citep{muja2009fast} used in GT-SLAM when dealing with ground texture images. In the optimization stage, GT-SLAM only constructs a simple pose graph, while ORB-SLAM3 optimizes a co-visibility graph that contains both poses and map points, which may result in the accuracy difference between these two systems.

\paragraph{Performance on the HD Ground Dataset} \label{sec:vo_hd_e}

The HD Ground dataset collects multiple subsequences for each ground texture. Due to the small image overlap and lack of rich textures, different methods may lose track on different subsequences, making it difficult to evaluate their performance on a specific ground texture using the average RMSE.
Therefore, we evaluate performance using the success rate, defined as $S_m = n_s / N_s$, instead of tracking error, where $N_s$ is the total number of subsequences and $n_s$ is the number of successful runs. A run on a subsequence is considered successful if the translational and rotational errors remain below $0.05 \meter$ and $10^{\circ}$, respectively.

The results are shown in \tref{tab:hd_ground_rate}. Due to insufficient overlap between adjacent images, SVO and ORB-SLAM3 fail to initialize on most sequences and are therefore omitted from the table.  
GT-SLAM$^{\text{w/o cvm}}$ denotes GT-SLAM without the constant velocity model for handling tracking loss.  
Our GroundSLAM achieves the highest success rate on 8 out of 11 textures and is the only system exceeding 70\% across all textures, demonstrating its robustness.

Feature-based methods perform well on textures with distinct corners, such as ``footpath" and ``workroom\_linoleum" sequences. However, they struggle on textures with few corners, like ``kitchen", or those with repetitive patterns, such as `terrace\_pavement" and ``checker\_plate\_steel", leading to frequent tracking failures. The VO experiment results on the HD Ground dataset align with the data association findings in \sref{sec:data_association_e}.  
The performance gap between GT-SLAM and GT-SLAM$^{\text{w/o cvm}}$ stems from GT-SLAM’s use of the constant velocity model for pose estimation during tracking loss. This allows it to complete successful runs even when losing track frequently on straight-path sequences.

\begin{table}[t]
    \caption{\centering Recall of map reuse(\%).}
    \label{tab:map_reuse}
    \centering
    \footnotesize
    \begin{tabular}{C{0.3\linewidth}C{0.15\linewidth}C{0.06\linewidth}C{0.17\linewidth}}
        \toprule
         Texture              & GT-SLAM   & Ours  & \textit{Improvement}         \\ \midrule
        bathroom\_tiles & 93.9  & \textbf{98.3}  & \textit{+ 4.4} \\
        checker\_plate\_steel & 85.3  & \textbf{99.1} & \textit{+ 13.8}  \\ 
        doormat & 48.7  & \textbf{88.3} & \textit{+ 39.6}  \\ 
        footpath & 66.1  & \textbf{93.4} & \textit{+ 27.3}  \\ 
        garage\_concrete & 56.2  & \textbf{99.9} & \textit{+ 43.7}  \\ 
        kitchen & 2.1  & \textbf{91.2} & \textit{+ 89.1}  \\ 
        office & 88.1  & \textbf{98.9} & \textit{+ 10.8}  \\ 
        parking\_place & 87.4  & \textbf{98.6} & \textit{+ 11.2}  \\ 
        ramp\_rubber & 21.6  & \textbf{94.2} &  \textit{+ 72.6}   \\ 
        terrace\_pavement & 61.5  & \textbf{92.8} & \textit{+ 31.3}  \\ 
        workroom\_linoleum & 90.9 & \textbf{99.8} & \textit{+ 8.9} \\
        \midrule
        average            & 63.8 & \textbf{95.9} & \textit{+ 32.1} \\
        \bottomrule     
    \end{tabular}
    \vspace{-1.0em}
\end{table}

\begin{table}[t]
    \footnotesize
    \caption{The RMSE (cm) comparison in the ablation study. The best results are \textbf{highlighted}. $\times$ refers to tracking lost}
    \label{tab:ablation}
    \centering
    \begin{threeparttable}
    \begin{tabular}{C{0.27\linewidth}|C{0.07\linewidth}C{0.07\linewidth}|C{0.09\linewidth}C{0.07\linewidth}C{0.11\linewidth}}
        \toprule
        \multirow{3}{*}{Sequence}  & \multicolumn{2}{c|}{Full Sequence}       & \multicolumn{3}{c}{Translation Parts Only\tnote{*}}   \\ 
          & \text{w/o k.}   & Ours  & w/o k.   & Ours     & \textit{Reducing}   \\ 
        \midrule
        Brick\_seq1           & $\times$ & \textbf{16.2}  & 2.4              & \textbf{1.8}  & \textcolor{red}{\textit{- 0.6}} \\
        Brick\_seq2           & $\times$ & \textbf{21.3}  & \textbf{3.0}     & 3.3           & \textcolor{blue}{\textit{+ 0.3}}  \\ 
        Carpet1\_seq1         & $\times$ & \textbf{110.7} & $\times$         & \textbf{8.9}  & \textcolor{red}{\textit{-}}  \\ 
        Carpet1\_seq2         & $\times$ & \textbf{42.3}  & 7.0              & \textbf{6.8}  & \textcolor{red}{\textit{- 0.2}}  \\ 
        Carpet2\_seq1         & $\times$ & \textbf{28.9}  & 4.1              & \textbf{3.7}  & \textcolor{red}{\textit{- 0.4}}  \\ 
        Carpet3\_seq1         & $\times$ & \textbf{17.4}  & $\times$         & \textbf{3.2}  & \textcolor{red}{\textit{-}}  \\ 
        Carpet3\_seq2         & $\times$ & \textbf{25.2}  & \textbf{3.0}     & 3.8           & \textcolor{blue}{\textit{+ 0.8}}  \\ 
        Carpet3\_seq3         & $\times$ & \textbf{37.6}  & \textbf{4.1}     & \textbf{4.1}  & \textcolor{red}{\textit{- 0.0}}  \\ 
        Coarse\_asphalt\_seq1 & $\times$ & \textbf{16.3}  & 3.5              & \textbf{2.6}  & \textcolor{red}{\textit{- 0.9}}   \\ 
        Concrete\_seq1        & $\times$ & \textbf{36.1}  & 20.9             & \textbf{3.2}  & \textcolor{red}{\textit{- 17.7}}  \\ 
        Concrete\_seq2        & $\times$ & \textbf{26.2}  & 3.9              & \textbf{3.8}  & \textcolor{red}{\textit{- 0.1}} \\
        Fine\_asphalt\_seq1   & $\times$ & \textbf{15.6}  & \textbf{2.7}     & 2.8           & \textcolor{blue}{\textit{+ 0.1}} \\
        Fine\_asphalt\_seq2   & $\times$ & \textbf{17.1}  & \textbf{2.4}     & 3.8           & \textcolor{blue}{\textit{+ 1.4}} \\
        Granite\_tiles\_seq1  & $\times$ & \textbf{46.9}  & $\times$         & \textbf{3.2}  & \textcolor{red}{\textit{-}} \\
        Granite\_tiles\_seq2  & $\times$ & \textbf{51.8}  & 3.8              & \textbf{1.4}  & \textcolor{red}{\textit{- 2.4}} \\
        Gravel\_road1\_seq1   & $\times$ & \textbf{16.6}  & 11.1             & \textbf{2.6}  & \textcolor{red}{\textit{- 8.5}} \\
        Gravel\_road2\_seq1   & $\times$ & \textbf{33.9}  & $\times$         & \textbf{3.7}  & \textcolor{red}{\textit{-}} \\
        \midrule
        average               & $\times$ & \textbf{32.9}  & 5.2\tnote{\dag}             & \textbf{3.7} & \textcolor{red}{\textit{- 2.2}}\tnote{\dag} \\
        \bottomrule     
    \end{tabular}
        \begin{tablenotes}
            \footnotesize
            \item[*] As the system without kernel (w/o k.) fails to estimate the rotation, we use the translational parts of the PathTex dataset for comparison.
            \item[\dag] The average value of the successful runs.
        \end{tablenotes}
    \end{threeparttable}
    % \vspace{-0.5em}
\end{table}

\begin{figure}[t]
    \vspace{0.5em}
    \centering
    \setlength{\abovecaptionskip}{0.8em}
    \includegraphics[width=0.95\linewidth]{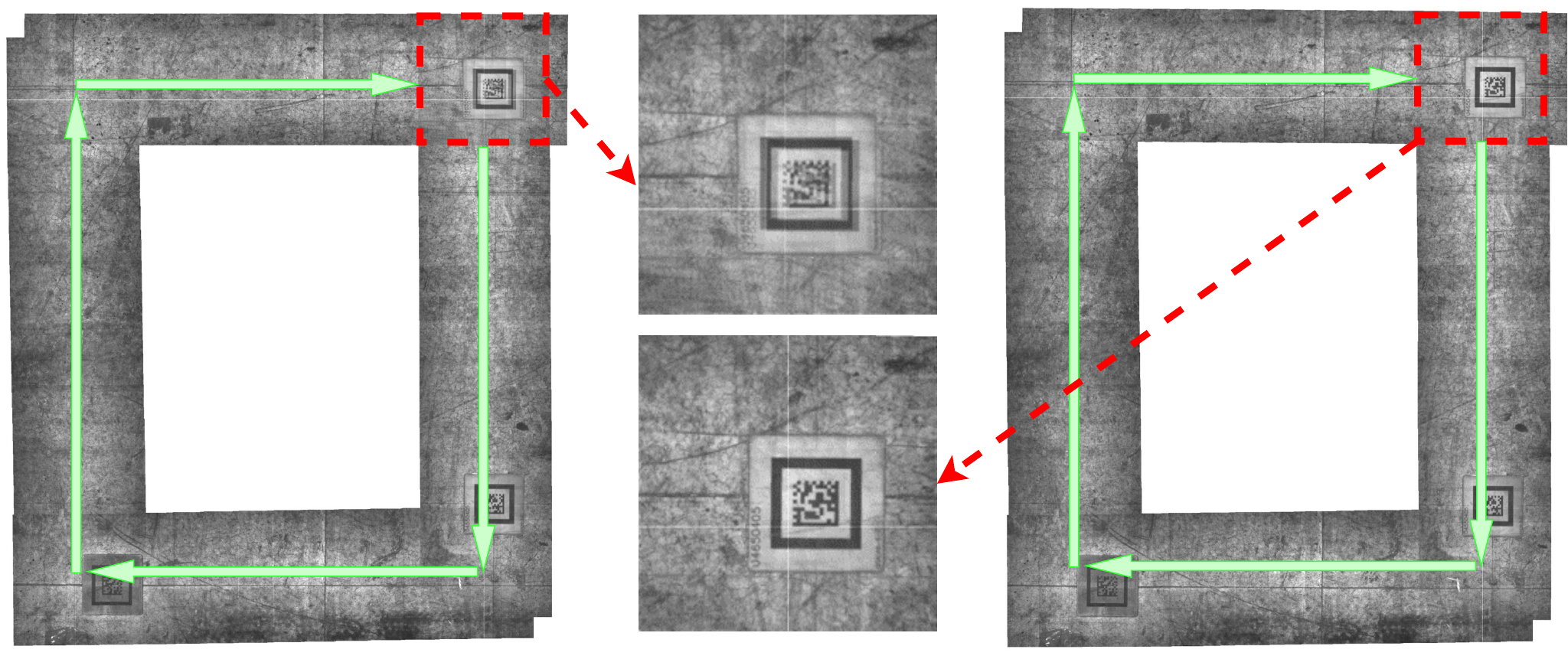}
    \caption{A live mapping demo of GroundSLAM without (left) and with (right) loop detection and correction. The robot moves in a rectangular path (the green line) with the starting and ending being at the same location (the red square). 
    The drift error of our VO module is only $\textbf{0.2\%}$ of the trajectory, which is significantly lower than other SOTA systems (about 1\%).
    The drift error causes the blurring in the stitched map (left), and it is eliminated by our loop correction module (right). The tag was not utilized for pose estimation but served solely as a reference ground truth.  }
    \label{fig:loop}
    % \vspace{-0.8em}
\end{figure}

\subsection{Loop Detection} \label{sec:loop_e}

Since the HD Ground dataset sequences do not contain loop closures, we evaluate our system's loop detection performance using only the PathTex dataset.
The results are summarized in \tref{tab:loop}, , where ``\text{w/o loop}" and ``\text{w/ loop}" represent the system without and with loop correction, respectively. ORB-SLAM3 is excluded due to frequent tracking failures across most sequences.
The results show that the introduced loop detection and correction techniques reduce the translational error by 30.2\%, 10.7\%, 0.6\%, and 23.5\% across the four sequences. Moreover, our system consistently outperforms GT-SLAM in all sequences and, even without loop closure correction, GroundSLAM achieves significantly better performance than GT-SLAM with loop closure correction, further demonstrating its robustness.
Compared with the results in \tref{tab:hd_ground_rate}, we find adding loop detection has increased the error of GT-SLAM on the sequence ``Carpet1\_seq2''. It is because GT-SLAM detected a false loop closure due to similar patterns. The incorrect loop detection did not appear in GroundSLAM, which indicates the robustness of our method.

\fref{fig:loop_trajectory} and \fref{fig:loop_error} show a comparison of our system with and without loop correction on the ``Fine\_asphalt\_seq2'' sequence. 
We controlled the robot to traverse three loops when collecting this sequence, ensuring sufficient loop closures.
It is seen that the trajectory of ``Ours \text{w/ loop}" is closer to the ground truth than the trajectory of ``Ours \text{w/o loop}". The pose errors are significantly decreased after the loop correction.
Note that the other methods such as SVO and ORB-SLAM3 constantly lose track in this sequence, which further indicates the robustness of our method.

\subsection{Map Reuse} \label{sec:map_reuse_e}

Map reuse enables drift-free localization using a prior map, and thus is essential for the real application. To compare our approach with feature-based methods, we select GT-SLAM, the state-of-the-art feature-based system, as the baseline. The HD Ground dataset is chosen for this evaluation since it contains both mapping and test data collected in the same areas. The mapping data serves as a database, while images from the test set are used as queries to retrieve similar database images. The pose of each query image is then estimated by computing its relative transformation to the matched database image. As detailed in \sref{sec:map_reuse}, a prior pose is provided for the query images to ensure robust relocalization, which is very common in real applications.

\begin{figure}[t]
    \vspace{0.5em}
    \centering
    \setlength{\abovecaptionskip}{0.5em}
    \includegraphics[width=0.99\linewidth]{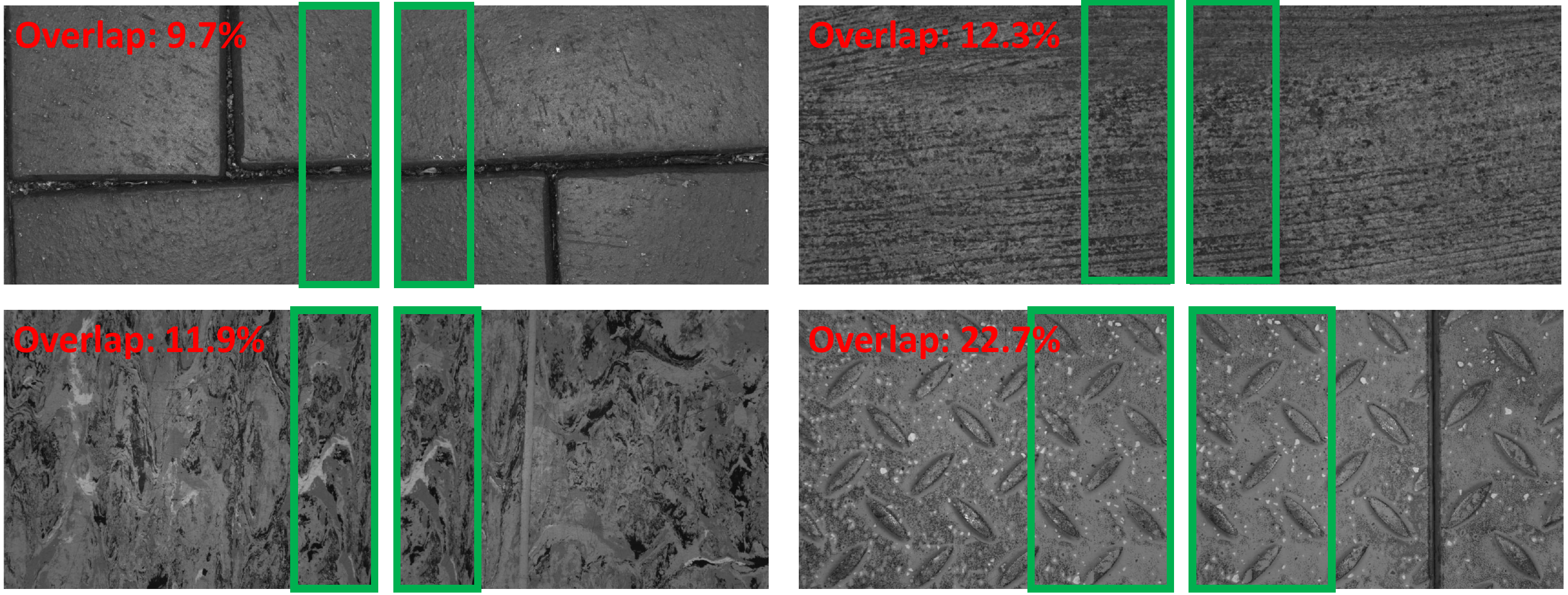}
    \caption{Failure cases of our system on the HD Ground dataset. Since the publicly available sequences are sampled from the original recordings, some adjacent images exhibit minimal overlap, leading to pose estimation challenges.}
    \label{fig:failure_case}
    \vspace{-0.8em}
\end{figure}

We generate the prior pose by randomly perturbing the ground truth position of each query image within a 0.5m radius and provide the initial pose to both GT-SLAM and our system.  
Localization is considered successful if the estimated pose has a translation error of less than 2mm (approximately 20 pixels) and a rotation error below $1.15^{\circ}$ (around 0.02 rad). Compared to the evaluation criteria in \sref{sec:vo_e}, this benchmark is significantly more stringent due to the drift-free nature of pose estimation in this context.  
To quantify performance, we introduce the recall rate, defined as the proportion of successfully localized query images relative to the total number of test images. The results, summarized in \tref{tab:map_reuse}, demonstrate that our system consistently outperforms GT-SLAM across various textures. Our system achieves recall rates exceeding 88\% in all cases, delivering an average improvement of 32.1\% over GT-SLAM. Notably, on textures with sparse corners, such as ``kitchen", the recall rate improvement approaches 90\%, highlighting the robustness and resilience of our approach.

\subsection{Live Mapping Demo} \label{sec:live-demo}

We present a real-time mapping demonstration incorporating loop detection and correction. The data was collected in a warehouse environment using an Automated Guided Vehicle (AGV), as depicted in \fref{fig:warehouse}. The AGV is equipped with a downward-facing camera, originally intended for localization via QR codes within the warehouse.  
As shown in \fref{fig:loop}, the robot followed a rectangular trajectory (highlighted in green), starting and stopping at the same location (marked by the red square), thereby forming a loop. Images along this path were then stitched using poses estimated by GroundSLAM, both without (left) and with (right) loop correction. Note that the tag was not utilized for pose estimation but served solely as a reference ground truth.  
Due to drift errors, the stitched map without loop correction exhibits noticeable blurring, which is effectively eliminated on the right. This clarity underscores the effective elimination of drift errors and the precise pose corrections.  
Additionally, we have tested GroundSLAM with this AGV in a real industrial environment spanning over 10,000 \square\meter. The system exhibited consistent robustness and accuracy.

\begin{figure}[t]
    \vspace{0.5em}
    \centering
    \setlength{\abovecaptionskip}{0.5em}
    \includegraphics[width=0.99\linewidth]{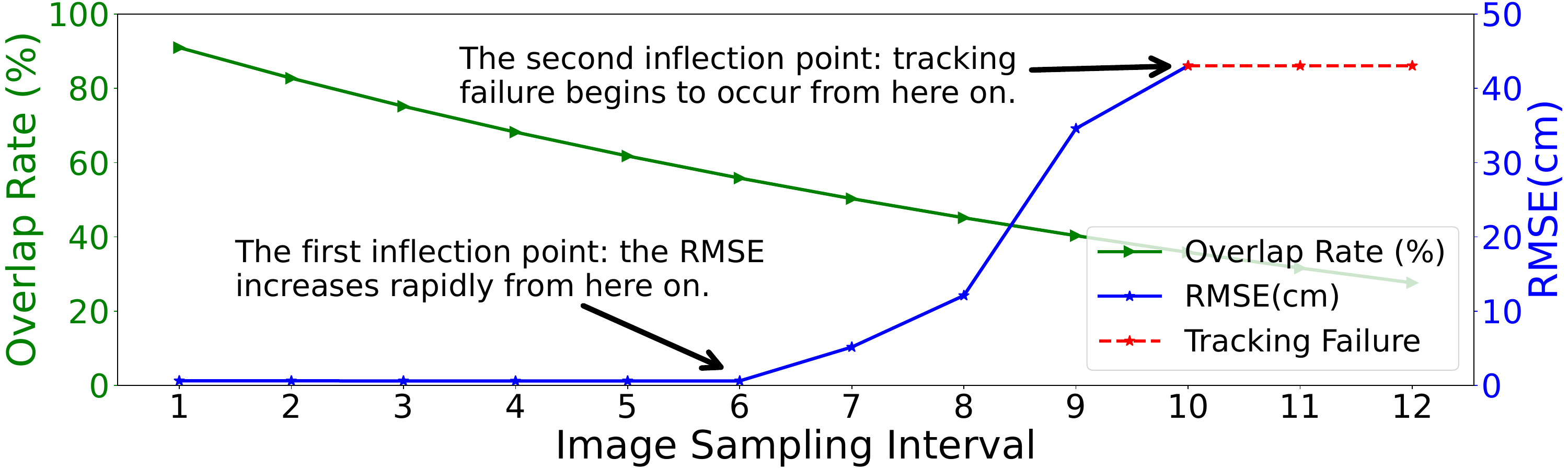}
    \caption{Average overlap rates and localization errors (RMSE) at different sampling intervals on the "Coarse\_asphalt" texture. The localization error increases significantly when the overlap rate drops below 50\%, and tracking failures start to occur when the overlap rate falls below 32\%.}
    \label{fig:overlap_rmse}
    % \vspace{-0.3em}
\end{figure}

\begin{figure}[t]
    % \vspace{0.5em}
    \centering
    \setlength{\abovecaptionskip}{0.5em}
    \includegraphics[width=0.99\linewidth]{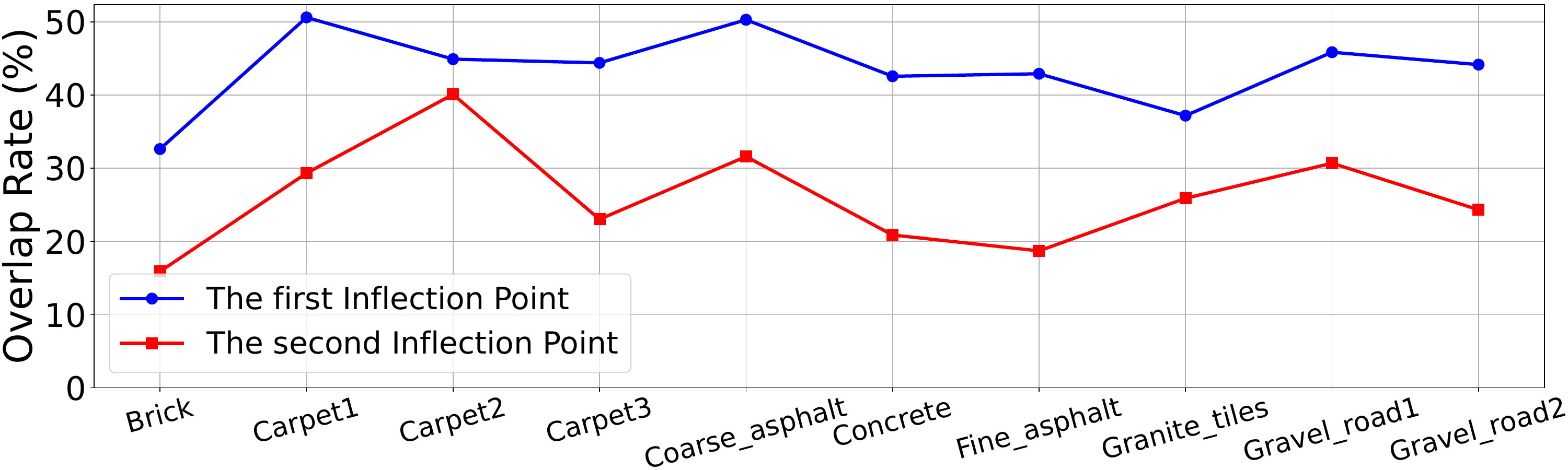}
    % \caption{The overlap rates of two inflection points for different ground textures. The average overlap rates of the first inflection point and the second inflection point are about 43\% and 25\%, respectively.}
    \caption{The overlap rates at two inflection points are analyzed across different ground textures. On average, the first inflection point occurs at an overlap rate of approximately 43\%, while the second occurs at around 25\%.}
    \label{fig:overlap_threshold}
    \vspace{-0.5em}
\end{figure}

\subsection{Ablation Study} \label{sec:ablation_study_e}

\paragraph{Kernel Method} \label{sec:ab_kernel}
The kernel method introduced in \sref{sec:methodology} plays a crucial role in rotation estimation, and removing the kernel function from KCC significantly increases the likelihood of motion estimation failure.  
To quantitatively assess its impact, we first evaluate the system without the kernel function and observe that it loses track in all sequences of the PathTex dataset. Next, to examine the contribution of the kernel method to translation estimation, we isolate the translational components of the PathTex dataset and use them to compare the performance of GroundSLAM with and without the kernel function. \tref{tab:ablation} presents the results, where ``w/o k.'' represents GroundSLAM without the kernel function.   
The results show that the incorporation of the kernel function reduces translation errors in 12 of 17 sequences, with an average error reduction of approximately 42\%. Furthermore, removing the kernel function leads to tracking failures in 4 sequences. These experiment results highlight the substantial benefits of the proposed kernel method: (1) it enhances the system's ability to estimate rotational motion, and (2) it improves both the accuracy and robustness of translation estimation.

\paragraph{Failure Case Analysis} \label{sec:ab_failure}

In this part, we provide an in-depth analysis of the failure cases of our system on the HD Ground dataset. Since the image sequences in the HD Ground dataset are sampled from original videos, they have a low frame rate, leading to minimal overlap between adjacent images. Examples of such cases are shown in \fref{fig:failure_case}. This insufficient overlap causes pose estimation failures of our system in \tref{tab:hd_ground_rate} and \tref{tab:map_reuse}.  
To further investigate the impact of the overlap rate on system performance, we apply uniform sampling to the original image sequences of the PathTex dataset, generating new sequences with varying overlap rates. \fref{fig:overlap_rmse} presents the average overlap rates and localization errors under different sampling intervals for the ``Coarse\_asphalt" texture. Two inflection points are observed: when the overlap rate drops below 50\%, localization error increases rapidly, and when it falls below 32\%, tracking failures begin to occur. Similar trends are observed across other textures.  
\fref{fig:overlap_threshold} summarizes the overlap rates at these inflection points for different ground textures, showing slight variations across textures. On average, the first and second inflection points occur at 43\% and 25\% overlap, respectively. Based on these findings, we conclude that for high-precision performance, GroundSLAM requires an overlap rate greater than 43\% between adjacent images, while maintaining an overlap above 25\% is necessary to prevent tracking failures.  

In our real-world application, our warehouse robot operates at a maximum speed of 2.5\meter/\second, ensuring a minimum image overlap of 53\%. As a result, our system can provide reliable localization. To deploy our system on a faster-moving robot, a higher-frame-rate camera or a higher-mounted camera with a wider field of view can be used to ensure the sufficient overlap between adjacent frames.

\begin{figure}[t]
    % \vspace{0.6em}
    \centering
    \setlength{\abovecaptionskip}{0.8em}
    \includegraphics[width=0.99\linewidth]{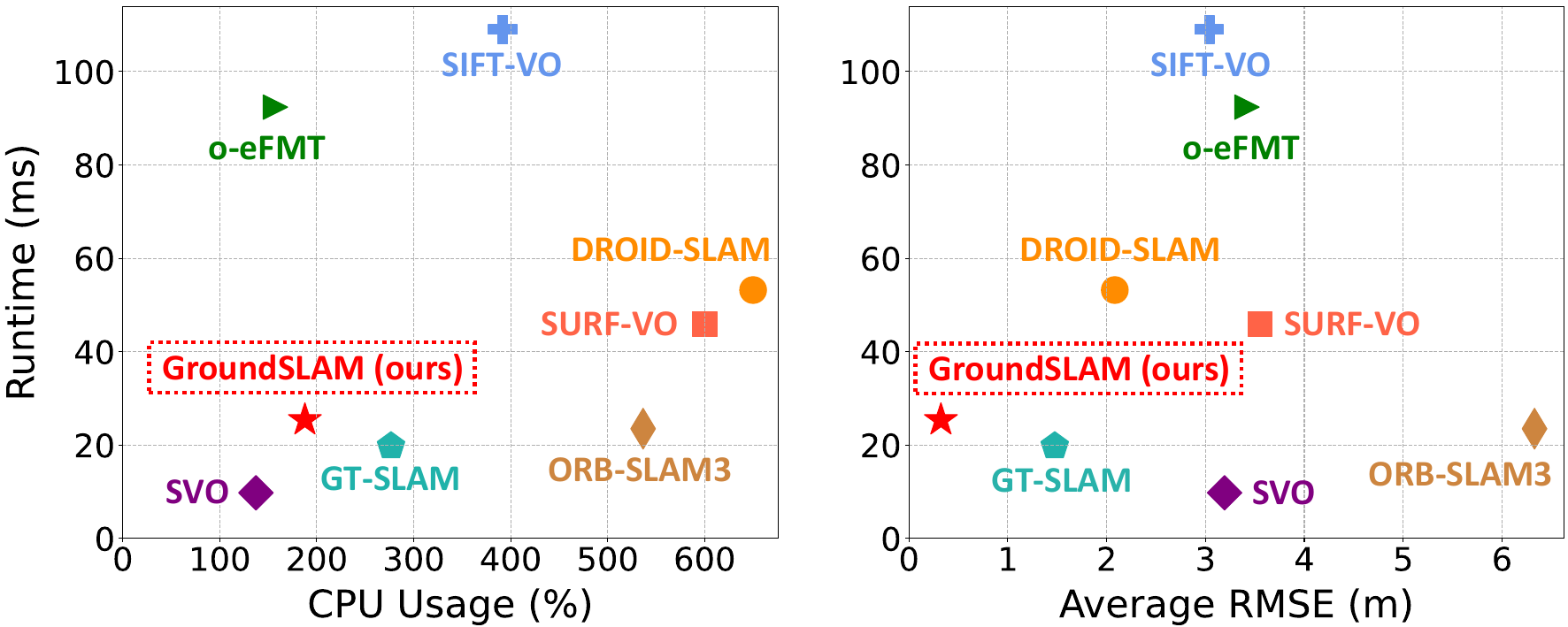}
    \caption{Efficiency analysis of different systems on the PathTex dataset. The vertical axis represents the average runtime per frame, while the horizontal axes correspond to CPU usage (left) and average RMSE (right), respectively.}
    \label{fig:vo_runtime}  
    \vspace{-0.4em}
\end{figure}

\subsection{Efficiency Analysis} \label{sec:efficiency_analysis_e}

This section presents the efficiency analysis of our GroundSLAM system. The evaluation is conducted on a computer with an Intel Core i9 CPU. We use the PathTex dataset and the image resolution is $640 \times 480$.
We compare the running speed, computing resource consumption, and the accuracy of different systems. The metrics are per-frame runtime, CPU usage, and average RMSE. For sequences where tracking failures occur, we assign an error of \( 10 \) meters when computing the average RMSE. The results are presented in \fref{fig:vo_runtime}.  
Overall, our system achieves the highest accuracy while maintaining real-time performance with minimal computational resource consumption. GroundSLAM demonstrates a runtime comparable to GT-SLAM and ORB-SLAM3 while utilizing fewer CPU resources. Although o-eFMT is also a correction-based ground-texture localization system, it exhibits significantly lower efficiency and accuracy compared to our approach, further validating the effectiveness of the proposed KCC.  
It is also worth noting that DROID-SLAM actually utilizes more than six CPU cores, and its CPU usage, and its CPU usage in \fref{fig:vo_runtime} is only for a compact presentation.

We also list the runtime of each module of our system in \tref{tab:ni_runtime}. The front-end processing for one frame takes 24.5 \milli\second~and the back-end modules take 33 \milli\second. 
Note that the back-end module is not executed on every frame: loop detection runs only on keyframes, and loop correction is performed only when a loop closure is successfully detected. Therefore, their impact on real-time performance is minimal.

\begin{table}[t]
    \small
    \caption{\centering Runtime analysis of each module in GroundSLAM.}
    \label{tab:ni_runtime}
    \centering
    \begin{tabular}{C{0.15\linewidth}|C{0.3\linewidth}C{0.35\linewidth}}
        \toprule
        & Module           & Average Runtime (\milli\second) \\ \midrule
        \multirow{6}{*}{Front-end} & 
          Undistortion          & 0.3            \\ 
        & FFT                   & 0.7             \\ 
        & Polar-Image           & 3.4             \\
        & Rotation              & 10.0            \\
        & Translation           & 8.7            \\
        & Keyframe Selection    & 0.1            \\
        & Others                & 1.3            \\
        % \midrule
        & Total                   & 24.5           \\
        \specialrule{0em}{1pt}{1pt}
        \hline   
        \specialrule{0em}{1pt}{1pt}
        \multirow{2}{*}{Back-end} &
          Loop Detection        & 22.3           \\
        & Loop Correction       & 10.7            \\
        \bottomrule
    \end{tabular}
    \vspace{0.6em}
\end{table}

\section{Conclusions}\label{sec:conclusion}

In this work, we introduced GroundSLAM, a novel visual SLAM system designed for warehouse robots equipped with a downward-facing camera, leveraging ground textures for localization. 
GroundSLAM integrates feature-free visual odometry, loop closure detection, and map reuse to achieve robust and efficient pose estimation.  
Specifically, we incorporated a kernel cross-correlator (KCC) for image-level matching, enhancing robustness and reliability when processing ground images with sparse features or repetitive patterns.
By relying solely on ground textures, GroundSLAM can provide reliable localization even in environments with dynamic objects and open spaces, where traditional localization systems using forward-facing cameras or LiDAR often struggle.
Extensive evaluations demonstrated the effectiveness of GroundSLAM, showcasing its advantages over SOTA ground-texture-based localization and visual SLAM systems. To further support research in this area, we have released both the source code and the PathTex dataset, which is the first ground-texture-based dataset with high-precision pose ground truth.

\begin{acks}
The authors would like to thank Dr. Yuan Shenghai and Geekplus Technology Co., Ltd. for collecting the data.
\end{acks}

{
% \small
\balance
\bibliographystyle{SageH}
\bibliography{
references/chenw,
references/kcc}
}

\appendix
\renewcommand{\thesection}{Appendix \Alph{section}}
\section{Cross-Correlation Theorem}\label{app:appendix-cross-correlation}

Cross-correlation is a similarity measurement of two signals as a function of the displacement of one relative to the other, which is also known as a sliding inner-product \citep{tahmasebi2012multiple}. Denote two real finite discrete signals as $\mathbf{x}, \mathbf{h} \in \mathbb{R}^{m}$, the circular cross-correlation $\otimes$ can be defined as
\begin{equation}\label{eq:appendix-cross-correlation}
	({\mathbf{x}} \otimes \mathbf{h})[m] = \sum_{i=0}^{n-1} \mathbf{x}[i] \mathbf{h}[(m+i)_{\bmod n}] = \mathbf{x}^T\mathbf{h}_{(m)},
\end{equation}
where the bracket $[m]$ is to access the $m$-th element of a vector, $\cdot_{\bmod n}$ is the modulo operator and the subscript $\cdot_{(m)}$ denotes the left circular rotation of a vector by $m$ elements.
The circular cross-correlation \eqref{eq:appendix-cross-correlation} is often calculated using discrete Fourier transform (DFT) based on the circular cross-correlation theorem, which is known as the cross-correlation theorem for short. 
We denote $\hat{\mathbf{x}} = \mathcal{F}(x)$ as the DFT output.
\begin{lemma}[Cross-Correlation Theorem \citep{bracewell1986fourier}]\label{thm:appendix-cross-correlation-theorem} The DFT of the circular cross-correlation \eqref{eq:appendix-cross-correlation} on two finite discrete signals is equivalent to element-wise conjugate multiplication of the DFT of individual signals, i.e.,
	\begin{equation}
		\mathcal{F}(\mathbf{x} \otimes \mathbf{h}) = \hat{\mathbf{x}} \odot \hat{\mathbf{h}}^*,
	\end{equation}
  where the superscript ${\cdot}^*$ denotes the complex conjugate and $\odot$ is the element-wise multiplication.
\end{lemma}

\thmref{thm:appendix-cross-correlation-theorem} indicates that a circular cross-correlation of two finite signals can be obtained via the element-wise product of their individual DFT. This is crucial for many applications, in which the DFT is often calculated by efficient fast Fourier transform (FFT). For example, Rader’s algorithm \citep{rader1968discrete} is only of complexity $\mathcal{O}(n\log n)$, where $n$ is the length of the signal.

\section{Boundary Effects}\label{app:appendix-boundary}
In the proof of \thmref{thm:kcc-equivariance}, we assume that the transformation $\mathcal{T}$ should be periodic. However, this isn't always a practical necessity. For the translation, the boundary effect can be diminished by applying a Gaussian mask to the image, as demonstrated by \cite{Henriques:2015jy}. Due to the periodic structure in the rotation space, \ie $\phi = \phi + 2 \pi$ where $\phi$ is a rotation angle, so the kernel vector $\boldsymbol{\kappa}_{\mathbf{z}}(\mathbf{x})$ is cyclic when it is extended to more than one circle. Hence, there is no boundary effect for the rotation correlator.
Different from the rotation, the periodic property
is not kept in the scale space. Therefore, the boundary effects may have a negative impact on the performance. It is possible to use the existing strategies to eliminate this effect \citep{Fernandez:2015fi, Galoogahi:2015hc}. One of the simplest methods is to add zero padding to the kernel vector: $\boldsymbol{\kappa}_{\mathbf{z}}(\mathbf{x}) = [\kappa(\mathbf{x}, \mathcal{T}_0(\mathbf{z})), \cdots, \kappa(\mathbf{x}, \mathcal{T}_{m-1}(\mathbf{z}), 0, \cdots, 0]^T$. In this case, the length of the kernel vector will be doubled.

\section{KCC Theoretical Significance}\label{app:appendix-kcf}

The theoretical contribution of KCC is substantial. Specifically, KCC is the first to (1) generalize translation to arbitrary transformations and (2) support arbitrary kernel functions. These advancements fundamentally overcome the theoretical limitations of prior approaches and open the door to a wide range of potential applications, such as SLAM in this work.

We next illustrate why prior approaches, such as KCF \citep{Henriques:2015jy}, are limited to translation and rely on kernel functions that treat all dimensions of the data uniformly, thereby highlighting the significance of KCC.
Given the training samples $\mathbf{x}_i\in \mathbb{R}^n$ and their targets $y_i\in \mathbb{R}$, the objective of KCF is to find a function $f(\mathbf{z}) = \mathbf{w}^\top \mathbf{z}$ that minimizes a squared error in the spatial domain:
\begin{equation}\label{eq:app_KCF_initial_obj}
  \min_{\mathbf{w}} \sum\limits_{i} \left( f(\mathbf{x}_i) - y_i \right)^2 + \lambda \|\mathbf{w}\|^2,
\end{equation}
where $\lambda$ is a hyperparameter that controls overfitting. The objective \eqref{eq:app_KCF_initial_obj} has a closed-form solution:
\begin{equation}\label{eq:app_KCF_initial_solution}
  \mathbf{w} = \left( \mathbf{X}^H \mathbf{X} + \lambda \mathbf{I} \right)^{-1} \mathbf{X}^H \mathbf{y},
\end{equation}
where the data matrix $\mathbf{X}$ is a row-stacked training samples of $\mathbf{x}_i$, and $\mathbf{y}$ is a stacked regression targets $y_i$, $\mathbf{X}^H$ denotes the Hermitian transpose of $\mathbf{X}$, and $\mathbf{I}$ is the identity matrix.
To avoid computing the inverse of the large matrix $\left( \mathbf{X}^H \mathbf{X} + \lambda \mathbf{I} \right)$, KCF assumes the training data samples are circulant:
\begin{equation}\label{eq:app_KCF_circulant}
    \mathbf{X} = \left[\begin{array}{ccccc} x_{1} & x_2 & x_3 & \cdots & x_{n} \\ 
    x_n & x_1 & x_2 & \cdots & x_{n-1} \\
     x_{n-1} & x_n & x_1 & \cdots & x_{n-2} \\
     \vdots & \vdots & \vdots & \ddots & \vdots \\
     x_2 & x_3 & x_4 & \cdots & x_1  \end{array}\right].
\end{equation}
This means that in KCF, a training sample $\mathbf{x}_i$ has to be a circular translation of any other training samples $\mathbf{x}_j$.
Based on the assumption, $\mathbf{X}$ can be computed from the DFT of the first row of $\mathbf{X}$, denoted as $\mathbf{x}$ \citep{gray2006toeplitz}:
\begin{equation}\label{eq:app_KCF_dft}
  \mathbf{X} = \mathbf{F} \text{diag} (\hat{\mathbf{x}}) \mathbf{F}^H,
\end{equation}
where $\mathbf{F}$ is a constant circulant matrix, and $\hat{\mathbf{x}}$ denotes the DFT of $\mathbf{x}$. Then $\mathbf{X}^H \mathbf{X}$ can be obtained through:
\begin{equation}\label{eq:app_xix}
  \begin{aligned}
    \mathbf{X}^H \mathbf{X} &= \mathbf{F} \text{diag} (\hat{\mathbf{x}}^*) \mathbf{F}^H \mathbf{F} \text{diag} (\hat{\mathbf{x}}) \mathbf{F}^H\\
    &= \mathbf{F} \text{diag} (\hat{\mathbf{x}}^*) \text{diag} (\hat{\mathbf{x}}) \mathbf{F}^H\\
    &= \mathbf{F} \text{diag} (\hat{\mathbf{x}}^* \odot \hat{\mathbf{x}}) \mathbf{F}^H,
  \end{aligned}
\end{equation}
where $\odot$ is the element-wise product. Substituting \eqref{eq:app_xix} into \eqref{eq:app_KCF_initial_solution} yields the solution in the frequency domain:
\begin{equation}\label{eq:app_w_fft}
  \begin{aligned}
    \hat{\mathbf{w}} &= \text{diag} \left( \frac{\hat{\mathbf{x}}^*}{\hat{\mathbf{x}}^* \odot \hat{\mathbf{x}} + \lambda} \right) \hat{\mathbf{y}}\\
    &=  \frac{\hat{\mathbf{x}}^* \odot \hat{\mathbf{y}}}{\hat{\mathbf{x}}^* \odot \hat{\mathbf{x}} + \lambda},
  \end{aligned}
\end{equation}
where the operator $\frac{\cdot}{\cdot}$ denotes the element-wise division. The solution in the spatial domain can be recovered with the inverse DFT.
By the techniques presented by \cite{rifkin2003regularized}, the solution in \eqref{eq:app_w_fft} can be reformulated as that of kernelized ridge regression for the nonlinear regression:
\begin{subequations}\label{eq:app_KCF_solution}
    \begin{align}
        \mathbf{w} &= \sum\limits_{i} \alpha_i \varphi(\mathbf{x}_i)  ,\\
        \mathbf{\alpha} &= (\mathbf{K} + \lambda \mathbf{I})^{-1} \mathbf{y},
    \end{align}
\end{subequations}
where $\varphi(\mathbf{x})$ maps the inputs of a linear problem to a non-linear feature-space.
$\mathbf{K}$ is the kernel matrix with elements $K_{ij} = \boldsymbol{\kappa}(\mathbf{x}_i, \mathbf{x}_j) = \varphi(\mathbf{x}_i)^\top \varphi(\mathbf{x}_j)$, $\boldsymbol{\kappa}$ denotes the kernel function, and coefficient $\alpha_i$ is the $i$-th element of $\boldsymbol{\alpha}$.

Therefore, since the solution in \eqref{eq:app_KCF_solution} depends on the assumptions in \eqref{eq:app_KCF_circulant} and \eqref{eq:app_KCF_dft}, the training data samples $\mathbf{x}_i$ are required to be \textit{mutually circulant}. Moreover, the kernel matrix $\mathbf{K}$ must preserve the circulant structure. KCF showed that the kernel matrix $\mathbf{K}$ is circulant only when the kernel function satisfies $\boldsymbol{\kappa}(\mathbf{x}_i, \mathbf{x}_j) = \boldsymbol{\kappa}(\mathbf{M}\mathbf{x}_i, \mathbf{M}\mathbf{x}_j)$ for any permutation matrix $\mathbf{M}$. This implies that the kernel function in KCF must treat all data dimensions equally.

Nevertheless, as defined in the objective function \eqref{eq:kernel-objective-any}, our KCC does not specify the transformation $\mathcal{T}_i$ and the kernel function $\kappa$.
This flexibility allows KCC to predict a wide range of affine transformations, including rotation, scale, and translation. As a result, we are able to design a robust and accurate visual SLAM system based on KCC.

\end{document}